\documentclass[electronics,article,accept,moreauthors,pdftex,10pt,a4paper]{Definitions/mdpi} 
\setitemize{parsep=6pt,itemsep=0pt,leftmargin=*,labelsep=5.5mm}
\setenumerate{parsep=6pt,itemsep=0pt,leftmargin=*,labelsep=5.5mm}
\setlist[description]{itemsep=0mm} 
\usepackage{amsmath,amssymb,amsfonts}
\usepackage{algorithmic}
\usepackage{graphicx}
\usepackage[utf8]{inputenc} 
\usepackage{textcomp}
\usepackage{hhline}
\usepackage{multirow}
\usepackage{tabularx}
\usepackage[table,xcdraw]{xcolor}
\usepackage{xcolor}
\usepackage{collcell}

\firstpage{1} 
\pubvolume{8}
\issuenum{1}
\articlenumber{57}
\pubyear{2019}
\copyrightyear{2019}
\history{Received: 27 November 2018; Accepted: 2 January 2019; Published: date}
\Title{Sign Language Representation by TEO Humanoid Robot: End-User Interest, Comprehension and Satisfaction}

\Author{Jennifer J. Gago *$^{,\dagger,\ddagger}$\orcidA{}, Juan G. Victores $^{\dagger,\ddagger}$\orcidB{} and Carlos Balaguer $^{\dagger}$\orcidC{}}

\AuthorNames{Jennifer J. Gago, Juan G. Victores and Carlos Balaguer}

\address[1]{%
  Robotics Lab, 
  University Carlos III de Madrid, 28911, Spain;
 jcgvicto@ing.uc3m.es (J.G.V.); balaguer@ing.uc3m.es (C.B.)}

\corres{\hangafter=1 \hangindent=1.05em \hspace{-0.82em}Correspondence: jenniferjoana.gago@uc3m.es or jennifer.gmunoz@gmail.com; Tel.: +34-91-624-6241}

\firstnote{\hangafter=1 \hangindent=1.05em \hspace{-0.82em} Current address: Av. de la Universidad, 30, 28911 Leganés, Madrid.} 
\secondnote{\hangafter=1 \hangindent=1.05em \hspace{-0.82em} These authors contributed equally to this work.}
\abstract{In this paper, we illustrate our work on improving the accessibility of Cyber--Physical Systems (CPS), presenting a study on human--robot interaction where the end-users are either deaf or hearing-impaired people. Current trends in robotic designs include devices with robotic arms and hands capable of performing manipulation and grasping tasks. This paper focuses on how these devices can be used for a different purpose, which is that of enabling robotic communication via sign language. For the study, several tests and questionnaires are run to check and measure how end-users feel about interpreting sign language represented by a humanoid robotic assistant as opposed to subtitles on a screen. Stemming from this dichotomy, dactylology, basic vocabulary representation and end-user satisfaction are the main topics covered by a delivered form, in which additional commentaries are valued and taken into consideration for further decision taking regarding robot-human interaction. The experiments were performed using TEO, a household companion humanoid robot developed at the University Carlos III de Madrid (UC3M), via representations in Spanish Sign Language (LSE), and a total of 16 deaf and hearing-impaired participants.}


\begin{document}
%%%%%%%%%%%%%%%%%%%%%%%%%%%%%%%%%%%%%%%%%%
%% Only for the journal Gels: Please place the Experimental Section after the Conclusions

%%%%%%%%%%%%%%%%%%%%%%%%%%%%%%%%%%%%%%%%%%
\setcounter{section}{0} 

\section{Introduction}
User accessibility and Universal Design (UD, also known as Design For All), are currently getting a growing consideration worldwide to reduce the physical and attitudinal barriers among people of all ages and abilities \cite{UD}. Regarding deaf and hearing-impaired people accessibility, Spanish and Catalan Sign Languages were recognised to be official languages in Spain by the national Parliament (BOE 27/2007) in 2007 \cite{b0}.

Several measures regarding the learning of this language from an early age and empowering deaf people to request interpreters in public and private services and areas have been taken. Following this approach, the use of resources that enhance and enable oral communication, such as lipreading, hearing aids, subtitling and other technological advances, has been declared a fundamental right. These measures aim to overcome any kind of discrimination of people with hearing disabilities in their access to information and communication, keeping in mind their heterogeneity and the specific needs of each group.

Regarding UD, there is a need to focus on the development of products that are easily accessible to as many people as possible, without the need to adapt or redesign them in a special way. In order to meet these objectives in the field of Cyber--Physical Systems (CPS), human--robot interaction must be not only accessible, but also usable. This guarantees easy access attributes and the possibility of understanding and learning how to communicate with the robot in a natural and intuitive way, without the need to investigate or get additional assistance.

Finding a way to make sure deaf or hearing-impaired individuals feel comfortable about interacting with technology is a step forward towards achieving the accessibility goal. The most widely used resource is to display subtitles on a screen, since sign language interpretation is not always an available option and it represents numerous challenges regarding its correct use and implementation. For that reason, there are many open questions whether or not it is likely that sign language users feel comfortable interacting with a robot in their everyday language. 

\subsection{Challenges of Representing Sign Language}

Representing sign language is a complex task which needs from advanced software and hardware to be done properly. It is not only a matter of precision, speed and movement fluidity, it is important to consider that signing is commonly complemented with facial expressions, shoulder raising, mouth morphemes, head tilt/nod/shake among other non-verbal communication signals that affect the meaning of the message, those are part of a set of behaviours called ``non-manual markers'' \cite{b4}.

The complexity of sign language is the main reason why it is still a quite incipient developing area in human--robot interaction, in comparison to other topics. There are relatively few projects related to robot reproduction of sign language. The assistant android developed in 2014 by Toshiba Corporation in collaboration with other Japanese technological institutes can mimic some simple movements, such as greetings and signing in Japanese \cite{bbc}. In addition, humanoids Robovie R3 (five-fingered robot) and Nao robot (three-fingered robot) were tested by the Istanbul Technical University for tutoring sign language in adults and children with typical hearing \cite{b1,b2}. This work proved the relevance of the hand anthropomorphism in sign language vocabulary comprehension. There are other studies regarding the design and development of robotic hands which have covered this topic independently from a humanoid robot, as it is the case of Project Aslan, from the University of Antwerp, which consists in a text dactylology translator arm \cite{aslan}.

Participatory Design (PD) has been considered, since involving users, designers and technology in a process of development and obtaining a distinct and diverse set of perspectives is highly valuable when developing a universal user oriented product \cite{CP}. It is important to take into consideration that the representation of sign language in CPS may be controversial without the feedback and participation of deaf and hearing-impaired people in the signing learning and implementation process. It is important to meet the expectations and needs of the target audience of this work before investing time and resources in specialising robots in certain areas. That is the main principle underlying this~project.

\subsection{TEO as a Household Companion}

TEO, also known as RH-2, is a full-size humanoid robot developed by researchers at the Robotics Lab research group, from UC3M. It features 28 Degrees of Freedom (DOF), two actuated hands and several sensors to provide it with information about its environment.

\begin{figure}[H]
  \centering
    \includegraphics[width=0.7\textwidth]{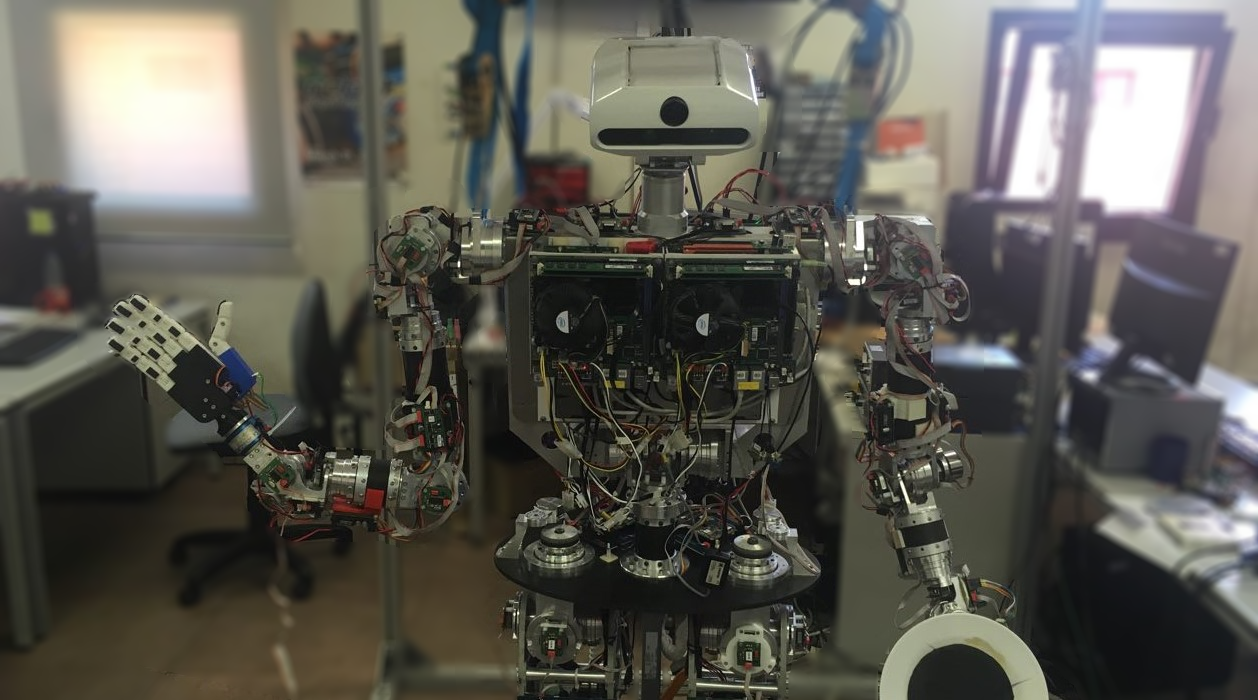}
  \caption{Humanoid robot TEO performing informal greeting with Dextra TPMG90-2 robotic hands.}
  \label{TEO}
\end{figure}

Regarding manipulation, TEO features two 6 DOF arms, each with a five-finger dexterous hand, which can be seen in Figure \ref{TEO}.
Thanks to their anthropomorphic characteristics, humanoid robots can perform human tasks such as greetings, waiter functions, folding and unfolding clothes, ironing and painting \cite{b3,b5}. Task performance is achieved by perception-manipulation loops  through a variety of machine learning techniques. 

As shown in the previous subsection, developing a robotic interpreter is an ambitious long-term project, since grammar, dialects, idioms and facial expression analyses would be needed. Currently, human--robot interaction with TEO relies on short command sentences delivered in both directions, so it is an affordable start point to test the user acceptance. To illustrate the interaction mentioned before, performing a greeting would consist in TEO using its voice to ask for the user name and, right after receiving that information, a short welcome sentence would be sent through the speakers. Therefore, the point of this work is to ensure this kind of communication can be established via sign language. 

\subsection{TEO Robotic Hands and Sign Language}

The development and adaption of new anthropomorphic robotic hands for TEO started in September 2017. Dextra TPMG90-2 is the version name of the current undergraduated hand prototypes operative and available in the robot \cite{b9}. They each have 15 DOF (14 for flexion/extension and 1 for abduction/adduction) and 6 actuators. The motion transmission system is based on a tendon-driven mechanism.

\begin{figure}[H]
  \centering
    \includegraphics[width=0.7\textwidth]{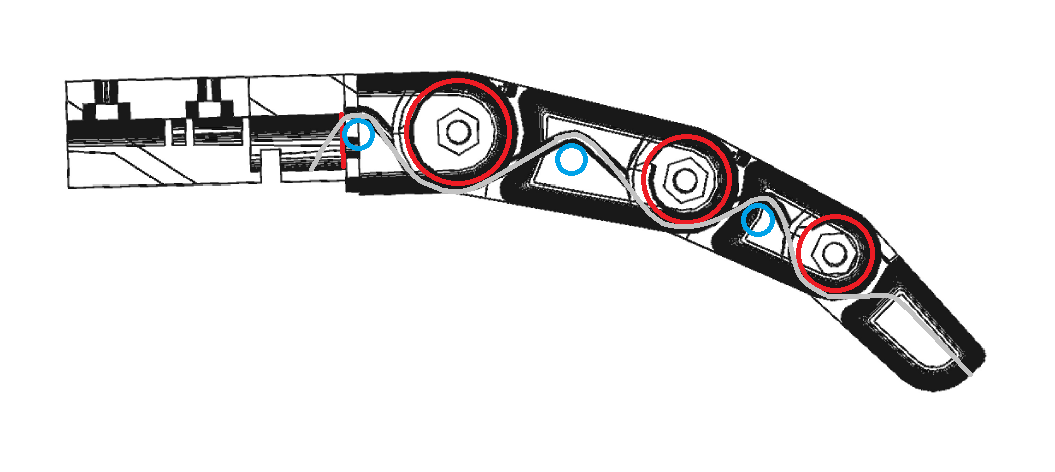}
  \caption{The phalange inner design allows the finger to show a natural gradual joint rotation starting from the proximal phalange.}
  \label{poleas}
\end{figure}

Underactuation could have been an issue regarding adaptability and precision, since each single actuator is in charge of flexing and extending all the phalanges of a single finger, with the exception of the thumb which is governed by two actuators. Contrary to this assumption, due to the phalange inner design depicted in Figure \ref{poleas}, the finger shows a natural gradual joint rotation that starts from the proximal phalange and allows the hand to develop movements similar to the one of the human hand.

\begin{figure}[H]
  \centering
    \includegraphics[width=0.97\textwidth]{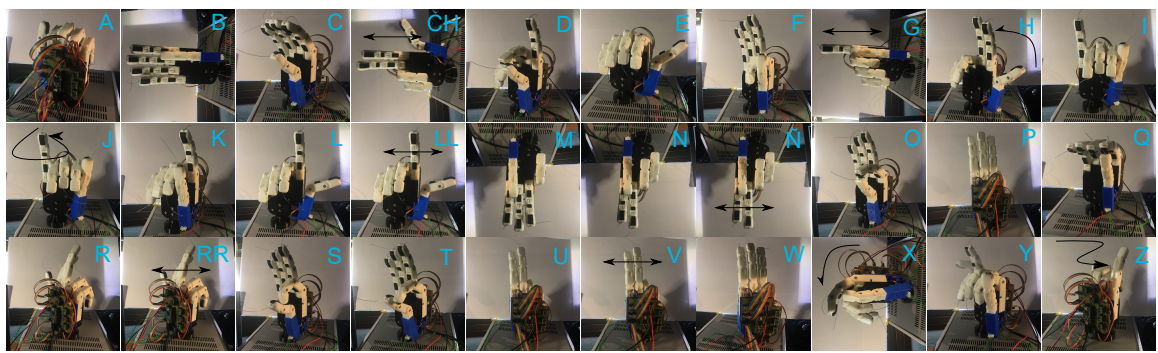}
  \caption{Spanish dactylology developed by Dextra TPMG90-2 robotic hand for joint position configuration. }
  \label{signs}
\end{figure}

Dactylology or fingerspelling requires a certain degree of position accuracy. Figure \ref{signs} shows how Dextra TPMG90-2 is able to represent the complete Spanish Sign Language (LSE) dactylology. This dactylology and its outcome demonstrate  how reasonable is to expect a positive performance in robotic hand signing. Since the hand is able to reproduce the complete alphabet, the following step is to test it with deaf and hearing-impaired users not related to the project to obtain and evaluate feedback.

\section{A Preliminary Study: Subtitles or Sign Language}

A general solution to procure deaf and hearing-impaired accessible communication in media and technology is to display subtitles. This settlement presents some advantages, such as ease of understanding, speed, or simplicity of implementation; and some disadvantages, such as the need of users' literacy, or the requirement of a sufficiently big readable screen. 

To measure the target audience preferences regarding human--robot interaction, a preliminary study is performed in this section to obtain the rate of users that prefer sign language over subtitles in this assistive robotics context, before and after watching a TEO humanoid robot demonstration. These~preferences are asked and shown as it is important to measure the user interest regarding the use of sign language within the context of humanoid robotics, before engaging in deeper studies.

\subsection{Preliminary Study Experimental Setup}

A group of 16 anonymous deaf and hearing-impaired users were recruited in collaboration with CILSEM (Spanish Sign Language Interpreters of Madrid Association) and Signapuntes Lengua de Signos (an LSE forum) and asked to choose between using sign language or subtitles with a humanoid robot, before and after watching a demonstration in which the robot asks ``how are you?'' in LSE. The sampling group consists of 16 Spanish men and women between 22 and 56 years old. The only characteristic taken into consideration for this study is the users' age, as the generational factor is considered to be the determining factor to measure users' predisposition to interact with or use technology. 

A statistical test is carried out to check the consistency in responses across the two options: sign language or subtitles. The same question is delivered on more than one occasion for each of the individuals included in the investigation, so the focus is on comparing whether the measurements made at two different times are the same or if, on the contrary, there is a significant change. McNemar's test \cite{b6a} fits perfectly for this purpose, since the data has one nominal variable with two categories and one independent variable with two connected groups, the sample is random, and sign language and subtitles are mutually exclusive \cite{b13}.

The importance of delivering this multiple choice test prior to the comprehension test needs to be highlighted. If most users refuse the idea of using LSE to interact with a robot in both cases, the~utility of the project should be reconsidered. Otherwise, if any or both of the cases receive a positive feedback, there would be sound arguments to continue with the research.

\subsection{Preliminary Study Results}

\begin{figure}[H]
  \centering
    \includegraphics[width=0.7\textwidth]{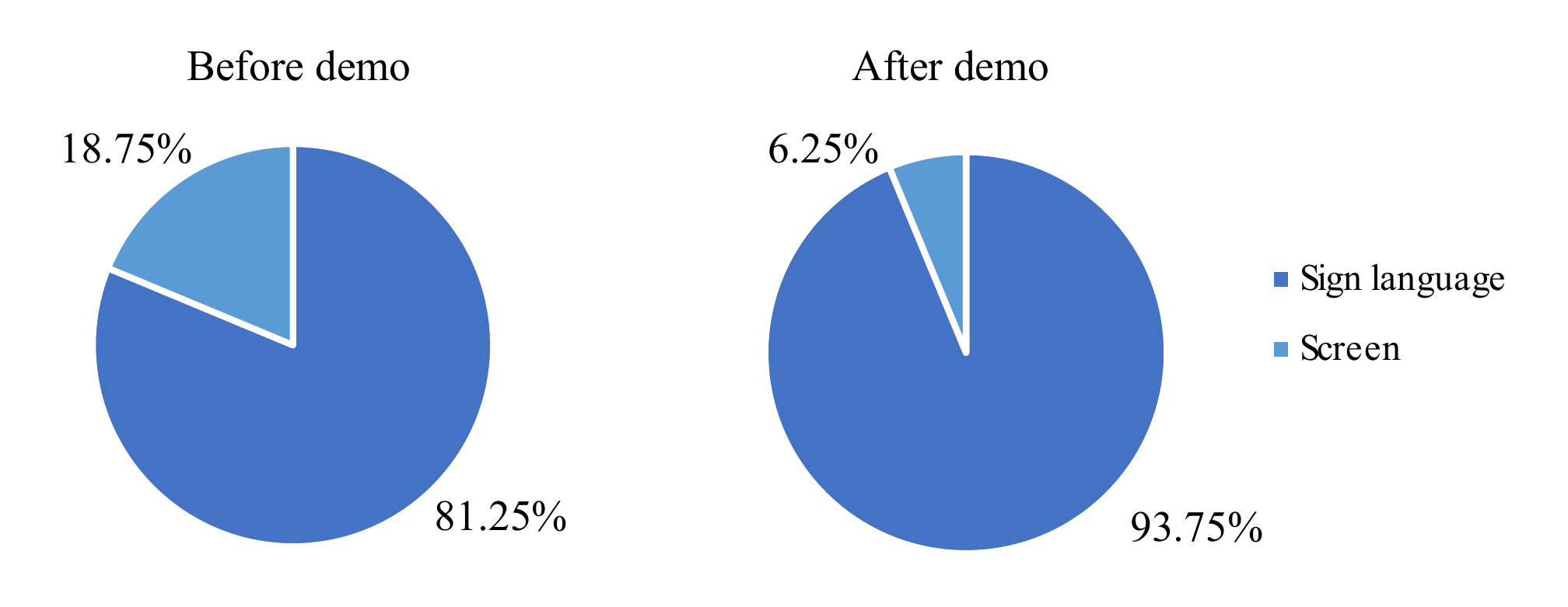}
  \caption{User preference rate regarding human--robot interaction before and after interacting with~TEO.}
  \label{subtitles}
\end{figure}

The experimental outcome is shown in Figure \ref{subtitles}. The user's predisposition to communicate with robots was over 80\% positive, and more than 65\% of reticent users changed their minds after their first experience with TEO. The experimental outcome predicts a positive response to human--robot interaction. However, a statistical analysis is needed to ensure this, which is performed in this section.

Table \ref{mcnemar} cluster the data, before and after demonstration, to analyse it via McNemar's test. If~there were no association between the results before and after demonstration, it is reasonable to expect the number of pairs where users before demonstration preferred sign language but users after demonstration did not (top right), to equal the number of pairs where the users after demonstration preferred sign language but the users before demonstration did not (bottom left). In this study, there were two discordant pairs (results before demonstration and results after demonstration had different exposure to the demonstration factor). There were a 100\% of pairs where users after demonstration preferred communicating via sign language but users before demonstration did not (bottom left), and no pairs where users before demonstration preferred communicating via sign language but users after demonstration did not (top right).

\begin{table}[H]
\centering
\caption{McNemar's test table that shows user preferences regarding human--robot interaction before and after the demonstration.}
\label{mcnemar}
\setlength{\tabcolsep}{3pt}
\begin{tabular}{ccccc}
\toprule
                                                            &                & \multicolumn{3}{c}{\textbf{After demonstration}} \\ \midrule 
                                                            &                & \textbf{Sign language} & \textbf{Subtitles} & \textbf{Total} \\ \midrule
\multicolumn{1}{c}{\multirow{4}{*}{\begin{tabular}[c]{@{}c@{}}\textbf{Before demonstration}\end{tabular}}} & \textbf{Sign language}     & 13         & 0          & 13             \\ \cmidrule{2-5} 
\multicolumn{1}{c}{}                                      & \textbf{Subtitles}     & 2          & 1          & 3              \\ \cmidrule{2-5} 
\multicolumn{1}{c}{}                                      & \textbf{Total} & 15         & 1          & 16             \\ \hline
\end{tabular}
\end{table}

Under the null hypothesis, with a sufficiently large number of discordants (elements of the antidiagonal), the chi-square ($\chi ^{2}$) test indicates that the distribution of the samples is chi-squared with 1 degree of freedom. 
\begin{equation}
\chi ^{2}={(b-c)^{2} \over b+c}
\end{equation}

When the elements of the antidiagonal sum less than 25, it is not well-approximated by the chi-squared distribution \cite{b6}. An alternative to the chi-squared distribution is the exact binomial test:
\begin{equation}
{\displaystyle {\text{exact-P-value}}=2\sum _{i=b}^{n}{n \choose i}0.5^{i}(1-0.5)^{n-i}}
\end{equation}

Edwards proposed a continuity corrected version of the McNemar test to approximate the binomial exact-P-value, which is the most widely used variant nowadays \cite{b7}:
\begin{equation}
    \chi ^{2}={(|b-c|-1)^{2} \over b+c}.
    \label{chi}
\end{equation}

From Equation (\ref{chi}), chi-squared equals 0.500 with 1 degrees of freedom. The P value is calculated with McNemar's test with the continuity correction and shows the probability of observing a large discrepancy between the number of the two kinds of discordant pairs. The two-tailed P value equals 0.4795. By conventional criteria, this difference is considered to be not statistically significant. Therefore, the percentage difference after and before watching TEO's demonstration can be attributed to chance and there is no consistent evidence of the effectiveness of TEO's performance in increasing the liking or interest rate. The odds ratio and its confidence interval cannot be calculated because one of the discordant values is zero. 

\section{Experimental Setup: Materials and Methods}

The first decision-making regarding the comprehension test setup is to consider how this test would be distributed. In order to preserve coherence in this experimental test, it is decided to keep using an anonymous online form distributed by LSE institutions and simulation-based multimedia files. There are several reasons for using simulation. On the one hand, this study aims to present the experiments in simulation as a first step within long-term work, where further studies will be performed with the physical humanoid robot. Therefore, the simulation outcome allows us to anticipate the effects of the embodiment and the robot appearance on user satisfaction and comprehension. On~the other hand, it is convenient to use a neutral background and a simplified representation of the humanoid robot that allows the respondents to focus on the gestures, since a non-neutral background could affect the quality of the gesture identification.

\begin{figure}[H]
  \centering
    \includegraphics[width=0.7\textwidth]{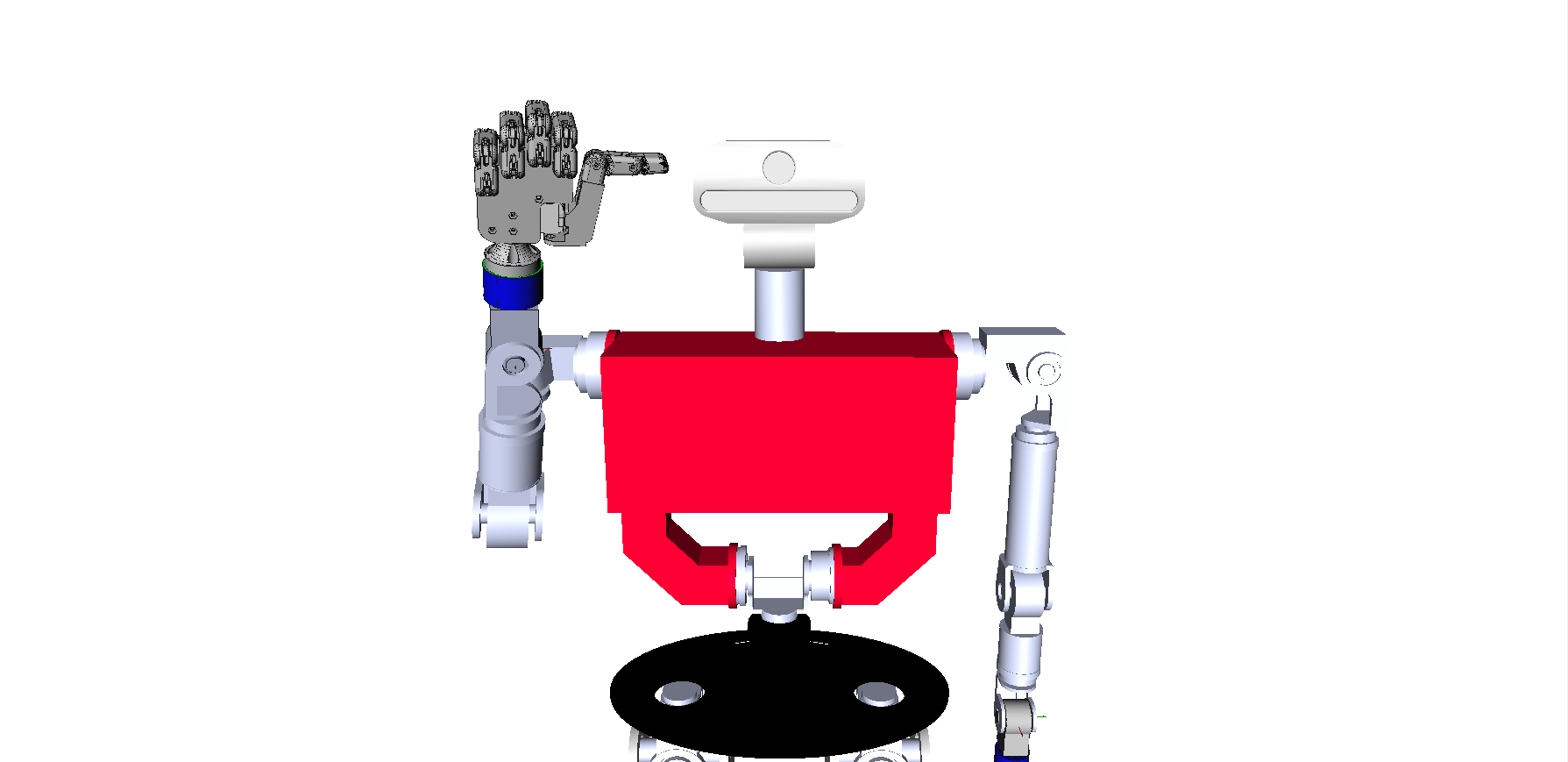}
  \caption{Frame of TEO's simulation signing letter E for the dactylology test. Vocabulary obtained from CNSE Foundation for the Suppression of Communication Barriers images and signs LSE database.}
  \label{letterE}
\end{figure}

TEO's signing simulation is developed by using OpenRAVE and QtCoin viewer, which provides a suitable environment for testing and developing. For that purpose, XML files were created to store all robot and scene descriptions. An example of this simulation can be found in Figure \ref{letterE}.

Usability testing is used to observe how easy to use sign language with TEO is by testing it with end-users. Participants are asked to complete these tests to detect problematic or confusing situations. Regarding the required number of participants to get acceptable results, Virzi \cite{b9a}, and~more recently Lewis \cite{b10} and Turner \cite{b11}, 
have published influential articles on the topic of sample size in usability testing. According to these authors, five is a proper number for usability testing, so~counting with 16~samples would be enough to develop a precise and reliable study and reach a successful conclusion~\cite{b12}.

The subjects of the test are randomly selected deaf and hearing-impaired subjects, contacted by CILSEM and Signapuntes Lengua de Signos. There is no detailed information given before the beginning of the test, and they are kindly asked to complete a form to obtain feedback about a signing humanoid robot. As commented before, the test is completely anonymous. The only personal information collected from the participants is their age, in order to detect any tendency regarding preferences or understanding.

The developed test consists in two main parts: dactylology and vocabulary recognition.  These two tests are selected to cover the study of the hand signing accuracy and the ability to communicate by using the upper part of the robot body. After the comprehension test with TEO, the user is ready to measure their satisfaction, so they will be asked to answer some questions about their experience.

Every test section is compulsory, which means that the responses cannot be submitted until the whole test is completed. There are just three additional optional questions about user preferences which can be completed at the end of each section.

\subsection{Dactylology}

Fingerspelling needs to be precise to be understood properly. There are some letters in LSE which share a quite similar hand configuration, so transitions, speed, and arm orientation must be treated carefully to obtain good results. It must be taken into account that TEO does not include anything similar to a human mouth, so it is not possible to aid the understanding of the signs with lip-speaking.

The confusion matrix of a class problem is a square matrix in which the columns are named according to the expected result, and the rows are named according to the experimental results. This~kind of matrix is the tool selected for showing explicitly when one letter is confused with another letter. It is a powerful tool since it allows to work separately with different types of errors.

The selected tool needs the provided test to check each one of the 30 letters of the Spanish alphabet. In order to avoid predictability and check if transitions between letters may cause any kind of confusion, the letters are shown in groups of three, so the user is asked to fill 10 blank gaps with 3~letters each. The letters are represented in a loop, so the first frame of each loop is marked with a blue dot to help the user to identify the beginning of the letter signing.

\subsection{Basic House Vocabulary}

The representation of sign language vocabulary involves the action of the upper body, which~includes hands, arms and head. This makes it specially important to coordinate all the simultaneous movements to make them seem human-like and, therefore, be more understandable by the end-user.

\begin{figure}[H]
  \centering
    \includegraphics[width=0.6\textwidth]{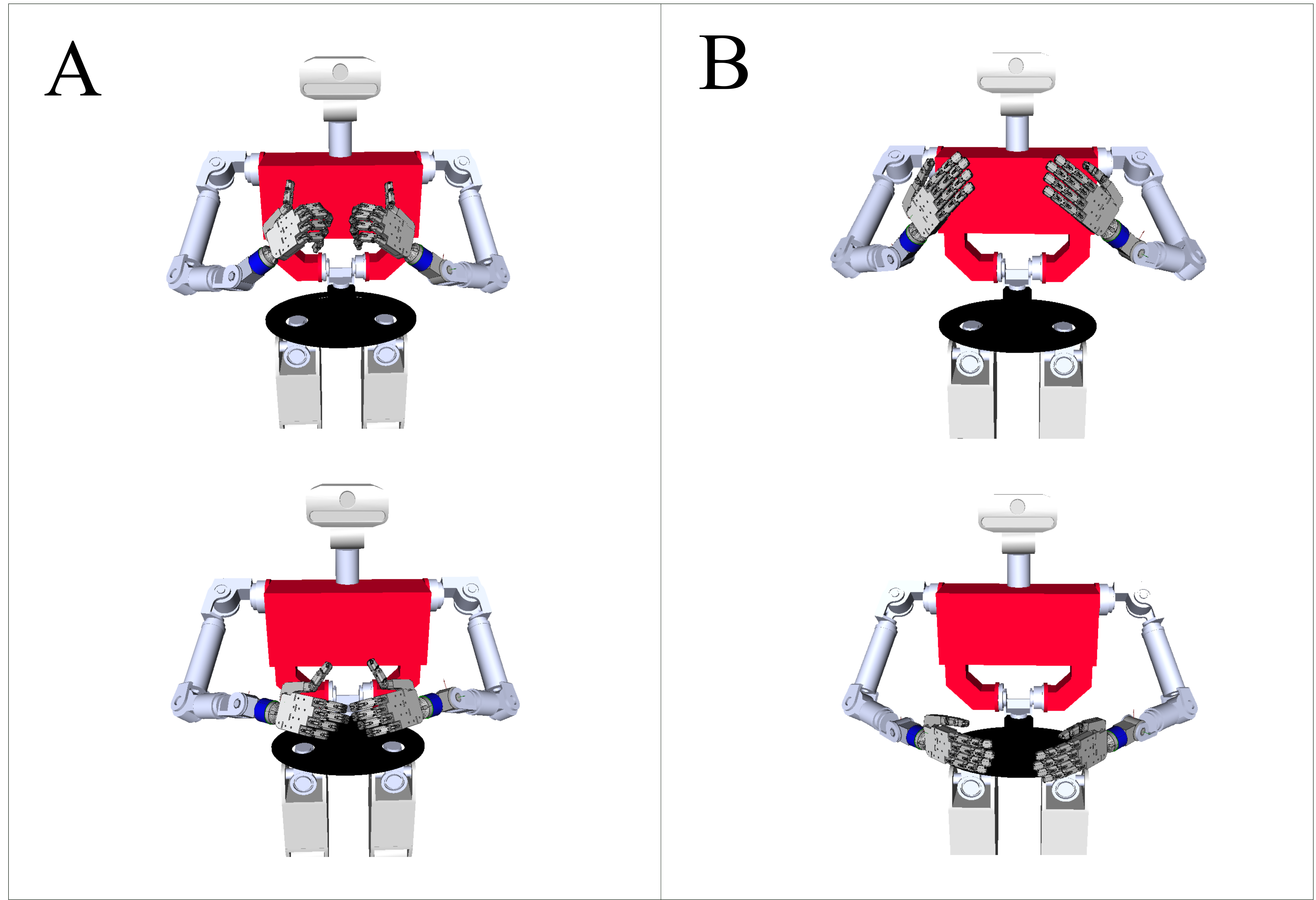}
  \caption{Initial and final frames of TEO's simulation signing (\textbf{A}) ``machine'' and (\textbf{B}) ``clothes'' for the vocabulary test. Vocabulary obtained from CNSE Foundation for the Suppression of Communication Barriers images and signs LSE database.}
  \label{paired}
\end{figure}

The tested vocabulary is selected according to the household companion context and considering some similar words to make it possible to apply the confusion matrix in this case, as well as in the dactylology test. There were nine related words and one unconnected word. ``Iron'' is the only unconnected word, selected due to its significance, since ironing is one of the most complex and relevant tasks that TEO can develop. ``Machine'' and ``clothes''; ``door'', ``kichen'' and ``closet''; ``bedroom'' and ``table''; and ``living room'' and ``telephone'' are the related words that are expected to lead to confusion. Figure \ref{paired} shows an example of the kind of similarity tested, where the arms' movement is quite similar in both cases, and the position of the fingers is fundamental to understand the difference in meaning.

In this case, as house vocabulary comprises a much wider group of words than the Spanish alphabet and to avoid obtaining unexpected results that could affect the confusion matrix and the following study, the users have to select the word from a ten choices drop-down list. Each word is shown independently, so, in accordance with the dactylology test, each user submits a ten-time~outcome.

\subsection{User Satisfaction}

An important part of this experimental test is to measure satisfaction, since it is fundamental that the end-users are not only able to communicate with the humanoid robot by using sign language, but that they can also do it in the most comfortable way. Six topics have been considered in order to measure user satisfaction, inspired in the users' overall valuation test developed for the ASIBOT assistive robot \cite{b12}, which are the following:

\begin{itemize}
    \item \textbf{Aesthetics}: Although TEO is still in an experimental phase and the way it currently looks is temporary. The outcome shows the way this topic affects the interaction experience.
    \item \textbf{Anthropomorphism}: The degree of anthropomorphism or human resemblance of the humanoid robot may influence the emergence of the uncanny valley \cite{b8}, so it must be taken into consideration.
    \item \textbf{Future prospects}: Since the technology shown in this test is under development, it is important to know if the user is willing or not to use it in the near future.
    \item \textbf{Comfort}: Uncomfortable experiences should not be present in assistive robotics, since these robots are made to work in close interaction with people; therefore, comfort must be handled properly. 
    \item \textbf{Comprehension ease}: The user may find some difficulties to comprehend the way TEO reproduces LSE which sometimes cannot be completely detected by error-proofing tests.
    \item \textbf{Usefulness}: Although preferences regarding robot communication are asked at the beginning of the form, end-users might consider human--robot interaction pointless after the tests.
\end{itemize}

The Likert scale is a measurement tool that, unlike binary questions that can be answered affirmatively or negatively, allows to measure attitudes and know the degree of conformity of the respondent with any proposed statement \cite{b13a}. It is especially appropriate in this context in which we want our end-users to provide their opinion quantitatively. In this sense, the response categories will serve to capture the intensity of the respondent's feelings toward each affirmation.

The most important requirement in this scale is that the distance between each possible answer choice is the same. It allows quantitative studies across different covered topics that have more than two outcome values \cite{b14}. There is no clear consensus among researchers about the number of response levels. The most commonly used scale consists in five levels; but four, seven, or ten levels are also frequently used~\cite{b15}. Adding levels results in obtaining more diverse valuations, as it avoids central tendency bias (CTB). CTB theory explains that in an item of only five levels, participants tend to avoid the two extreme options, obtaining very little variation.

The CTB effect could be softened by balancing positive and negative levels in the scale (symmetric scale) and letting the user respond to the test anonymously to avoid the pressure of being judged for selecting extreme options. A symmetric scale could also help to avoid acquiescence bias, which is a tendency of the respondent to agree or show positive feedback \cite{b17}. Since a five-level scale allows neutral response and two different levels of agreement and disagreement, which simplifies decision taking, it~is selected for the user satisfaction test. Table \ref{levels} shows the displayed options in the final survey.

\begin{table}[H]
\centering
\caption{Five-level Likert scale used in the user satisfaction questionnaire.}
\label{levels}
\setlength{\tabcolsep}{3pt}
\begin{tabular}{cc}
\toprule
\textbf{Scale Value} & \textbf{Opinion}           \\ \midrule
$-$2                   & Strongly disagree          \\ 
$-$1                   & Disagree                   \\ 
0                    & Neither agree nor disagree \\ 
1                    & Agree                      \\ 
2                    & Strongly agree             \\ \bottomrule
\end{tabular}
\end{table}

It is difficult to treat neutral responses, such as the ``neither agree nor disagree'' presented in the table, but it is recommended to offer the possibility of taking this option if the respondent is unsure about their opinion or  cannot decide between a positive or a negative answer. About considering a middle option as ``unsure'' or ``neutral'', a study developed by R. Amstrong found the differences to be imperceptible \cite{b16}.

\subsection{Optional Questions}
Some optional questions are provided in the delivered form at the end of each previous described test sections, to obtain further information regarding the respondent preferences. The three questions presented in the questionnaire are:

\begin{itemize}
    \item Regarding human--robot interaction, would you prefer any alternative method to using sign language or reading subtitles?
    \item Why do you prefer the way of interacting with TEO that you selected?
    \item What would you improve about TEO signing performance?
\end{itemize}

The answers to these questions could provide additional details that would help us to understand some issues that need to be fixed in future developments.

\section{Experimental Results}
\label{sec:guidelines}

Experimental results were collected two weeks after delivering the online form to the institutions involved in its distribution.
This limit on the period of time for receiving the form was established to assure distribution only within the reach of the target end-user group, as the link to the form was open and based on trust of anonymous user data. A total of 16 users participated up to that date.

\subsection{Dactylology}

Dactylology answers, provided the fact that robot movements were programmed by LSE non-experts, were surprisingly accurate and insightful. Table \ref{dactyl} depicts the confusion matrix that compares expected and obtained results. The elements in the main diagonal show the amount of correct answers for each specific letter. The `Other'' row contains the sum of answers that are not elements of the expected answers. It is noticeable at first sight that, except for the letters F and RR which will be commented within this section, all letters obtained a correct answer rate above 50\%. One third of the alphabet was completely understood (10 letters), with no failed attempts (discarding outlier answers). Finally, the mean shows that approximately 82\% of the answers were correct (369 correct answers over 450), which can be considered a successful outcome.

\begin{table}[H]

\centering
\small
\caption{Confusion matrix: dactylology.
The elements of diagonal, which represent correct answers, are marked in bold. Elements with a shaded background mean 100\% correct answers, discarding those of the ``Other'' row. In the Spanish alphabet, the CH, LL, RR and Ñ represent individual letters.}
\label{dactyl}
\setlength{\tabcolsep}{3pt}
\resizebox{\textwidth}{!}{
{\begin{tabular}{|c|c|c|c|c|c|c|c|c|c|c|c|c|c|c|c|c|c|c|c|c|c|c|c|c|c|c|c|c|c|c|c|}
\noalign{\hrule height 1.0pt}
                                                         & \textbf{}     & \multicolumn{30}{c|}{\textbf{Result Expected}}                                                                                                             \\
                                                          \cline{2-32} 
\textbf{}                                                &               & \textbf{A}                                                 & \textbf{B}                         & \textbf{C}                                                 & \textbf{CH}                                                & \textbf{D}                                                 & \textbf{E}                         & \textbf{F}                        & \textbf{G}                                                 & \textbf{H}                        & \textbf{I}                         & \textbf{J}                         & \textbf{K}                        & \textbf{L}                                                 & \textbf{LL}                        & \textbf{M}                                                 & \textbf{N}                                                 & \textbf{Ñ}                        & \textbf{O}                         & \textbf{P}                         & \textbf{Q}                                                 & \textbf{R}                                                 & \textbf{RR}                       & \textbf{S}                         & \textbf{T}                         & \textbf{U}                         & \textbf{V}                         & \textbf{W}                         & \textbf{X}                         & \textbf{Y}                         & \textbf{Z}                        \\ 
 \cline{2-32} 
\multicolumn{1}{|c|}{}                                   & \textbf{A}    & \color[HTML]{00629B} \textbf{14} & \color[HTML]{00629B} 1           & 0                                                          & 0                                                          & 0                                                          & 0                                  & 0                                 & 0                                                          & 0                                 & 0                                  & 0                                  & 0                                 & 0                                                          & 0                                  & 0                                                          & 0                                                          & 0                                 & 0                                  & 0                                  & 0                                                          & 0                                                          & 0                                 & 0                                  & 0                                  & 0                                  & 0                                  & 0                                  & 0                                  & 0                                  & 0                                 \\
 \cline{2-32} 
\multicolumn{1}{|c|}{}                                   & \textbf{B}    & 0                                                          & {\color[HTML]{00629B} \textbf{13}} & 0                                                          & 0                                                          & 0                                                          & 0                                  & 0                                 & 0                                                          & 0                                 & 0                                  & 0                                  & 0                                 & 0                                                          & 0                                  & 0                                                          & 0                                                          & 0                                 & 0                                  & 0                                  & 0                                                          & 0                                                          & 0                                 & 0                                  & 0                                  & 0                                  & 0                                  & 0                                  & 0                                  & 0                                  & 0                                 \\
 \cline{2-32} 
\multicolumn{1}{|c|}{}                                   & \textbf{C}    & 0                                                          & 0                                  & \color[HTML]{00629B} \textbf{15} & 0                                                          & 0                                                          & 0                                  & 0                                 & 0                                                          & 0                                 & 0                                  & 0                                  & 0                                 & 0                                                          & 0                                  & 0                                                          & 0                                                          & 0                                 & 0                                  & 0                                  & 0                                                          & 0                                                          & 0                                 & 0                                  & 0                                  & 0                                  & 0                                  & 0                                  & 0                                  & 0                                  & 0                                 \\ 
\cline{2-32} 
\multicolumn{1}{|c|}{}                                   & \textbf{CH}   & 0                                                          & 0                                  & 0                                                          &  {\color[HTML]{00629B} \textbf{15}} & 0                                                          & 0                                  & 0                                 & 0                                                          & {\color[HTML]{00629B} 4}          & 0                                  & 0                                  & {\color[HTML]{00629B} 1}          & 0                                                          & 0                                  & 0                                                          & 0                                                          & 0                                 & 0                                  & 0                                  & 0                                                          & 0                                                          & 0                                 & 0                                  & 0                                  & 0                                  & 0                                  & 0                                  & 0                                  & 0                                  & 0                                 \\ 
\cline{2-32} 
\multicolumn{1}{|c|}{}                                   & \textbf{D}    & 0                                                          & 0                                  & 0                                                          & 0                                                          &  {\color[HTML]{00629B} \textbf{15}} & 0                                  & 0                                 & 0                                                          & 0                                 & 0                                  & 0                                  & 0                                 & 0                                                          & 0                                  & 0                                                          & 0                                                          & 0                                 & 0                                  & 0                                  & 0                                                          & 0                                                          & {\color[HTML]{00629B} 2}          & {\color[HTML]{00629B} 1}           & 0                                  & 0                                  & 0                                  & 0                                  & 0                                  & 0                                  & 0                                 \\ \cline{2-32} 
\multicolumn{1}{|c|}{}                                   & \textbf{E}    & 0                                                          & 0                                  & 0                                                          & 0                                                          & 0                                                          & {\color[HTML]{00629B} \textbf{13}} & 0                                 & 0                                                          & 0                                 & 0                                  & 0                                  & 0                                 & 0                                                          & 0                                  & 0                                                          & 0                                                          & 0                                 & 0                                  & 0                                  & 0                                                          & 0                                                          & 0                                 & 0                                  & 0                                  & 0                                  & 0                                  & 0                                  & 0                                  & 0                                  & 0                                 \\
 \cline{2-32} 
\multicolumn{1}{|c|}{}                                   & \textbf{F}    & 0                                                          & 0                                  & 0                                                          & 0                                                          & 0                                                          & 0                                  & {\color[HTML]{00629B} \textbf{6}} & 0                                                          & 0                                 & 0                                  & 0                                  & 0                                 & 0                                                          & 0                                  & 0                                                          & 0                                                          & 0                                 & 0                                  & 0                                  & 0                                                          & 0                                                          & 0                                 & 0                                  & {\color[HTML]{00629B} 2}           & 0                                  & 0                                  & 0                                  & 0                                  & 0                                  & 0                                 \\
 \cline{2-32} 
\multicolumn{1}{|c|}{}                                   & \textbf{G}    & 0                                                          & 0                                  & 0                                                          & 0                                                          & 0                                                          & {\color[HTML]{00629B} 1}           & 0                                 &  {\color[HTML]{00629B} \textbf{15}} & {\color[HTML]{00629B} 1}          & 0                                  & 0                                  & 0                                 & 0                                                          & 0                                  & 0                                                          & 0                                                          & 0                                 & 0                                  & 0                                  & 0                                                          & 0                                                          & 0                                 & 0                                  & 0                                  & 0                                  & 0                                  & 0                                  & 0                                  & 0                                  & 0                                 \\
 \cline{2-32} 
\multicolumn{1}{|c|}{}                                   & \textbf{H}    & 0                                                          & 0                                  & 0                                                          & 0                                                          & 0                                                          & 0                                  & 0                                 & 0                                                          & {\color[HTML]{00629B} \textbf{9}} & 0                                  & 0                                  & {\color[HTML]{00629B} 3}          & 0                                                          & 0                                  & 0                                                          & 0                                                          & 0                                 & 0                                  & 0                                  & 0                                                          & 0                                                          & 0                                 & 0                                  & 0                                  & 0                                  & 0                                  & 0                                  & 0                                  & 0                                  & 0                                 \\ \cline{2-32} 
\multicolumn{1}{|c|}{}                                   & \textbf{I}    & 0                                                          & 0                                  & 0                                                          & 0                                                          & 0                                                          & 0                                  & 0                                 & 0                                                          & 0                                 & {\color[HTML]{00629B} \textbf{13}} & {\color[HTML]{00629B} 1}           & 0                                 & 0                                                          & 0                                  & 0                                                          & 0                                                          & 0                                 & 0                                  & 0                                  & 0                                                          & 0                                                          & 0                                 & 0                                  & 0                                  & 0                                  & 0                                  & 0                                  & 0                                  & {\color[HTML]{00629B} 2}           & {\color[HTML]{00629B} 1}          \\ \cline{2-32} 
\multicolumn{1}{|c|}{}                                   & \textbf{J}    & 0                                                          & {\color[HTML]{00629B} 1}           & 0                                                          & 0                                                          & 0                                                          & 0                                  & 0                                 & 0                                                          & 0                                 & 0                                  & {\color[HTML]{00629B} \textbf{12}} & 0                                 & 0                                                          & 0                                  & 0                                                          & 0                                                          & 0                                 & 0                                  & 0                                  & 0                                                          & 0                                                          & 0                                 & 0                                  & 0                                  & 0                                  & 0                                  & 0                                  & {\color[HTML]{00629B} 1}           & 0                                  & {\color[HTML]{00629B} 6}          \\ \cline{2-32} 
\multicolumn{1}{|c|}{}                                   & \textbf{K}    & 0                                                          & 0                                  & 0                                                          & 0                                                          & 0                                                          & 0                                  & 0                                 & 0                                                          & {\color[HTML]{00629B} 1}          & 0                                  & 0                                  & {\color[HTML]{00629B} \textbf{8}} & 0                                                          & 0                                  & 0                                                          & 0                                                          & 0                                 & 0                                  & 0                                  & 0                                                          & 0                                                          & 0                                 & 0                                  & 0                                  & 0                                  & 0                                  & 0                                  & 0                                  & 0                                  & 0                                 \\ 
\cline{2-32} 
\multicolumn{1}{|c|}{}                                   & \textbf{L}    & 0                                                          & 0                                  & 0                                                          & 0                                                          & 0                                                          & 0                                  & 0                                 & 0                                                          & 0                                 & {\color[HTML]{00629B} 1}           & 0                                  & 0                                 &  {\color[HTML]{00629B} \textbf{15}} & {\color[HTML]{00629B} 3}           & 0                                                          & 0                                                          & 0                                 & 0                                  & 0                                  & 0                                                          & 0                                                          & 0                                 & 0                                  & 0                                  & 0                                  & 0                                  & 0                                  & 0                                  & 0                                  & 0                                 \\ \cline{2-32} 
\multicolumn{1}{|c|}{}                                   & \textbf{LL}   & 0                                                          & 0                                  & 0                                                          & 0                                                          & 0                                                          & 0                                  & 0                                 & 0                                                          & 0                                 & 0                                  & 0                                  & 0                                 & 0                                                          & {\color[HTML]{00629B} \textbf{12}} & 0                                                          & 0                                                          & 0                                 & 0                                  & 0                                  & 0                                                          & 0                                                          & 0                                 & 0                                  & 0                                  & 0                                  & 0                                  & 0                                  & 0                                  & 0                                  & 0                                 \\ \cline{2-32} 
\multicolumn{1}{|c|}{}                                   & \textbf{M}    & 0                                                          & 0                                  & 0                                                          & 0                                                          & 0                                                          & 0                                  & 0                                 & 0                                                          & 0                                 & 0                                  & 0                                  & 0                                 & 0                                                          & 0                                  &  {\color[HTML]{00629B} \textbf{15}} & 0                                                          & 0                                 & 0                                  & {\color[HTML]{00629B} 1}           & 0                                                          & 0                                                          & 0                                 & 0                                  & 0                                  & 0                                  & 0                                  & {\color[HTML]{00629B} 1}           & 0                                  & 0                                  & 0                                 \\\cline{2-32} 
\multicolumn{1}{|c|}{}                                   & \textbf{N}    & 0                                                          & 0                                  & 0                                                          & 0                                                          & 0                                                          & 0                                  & 0                                 & 0                                                          & 0                                 & 0                                  & 0                                  & 0                                 & 0                                                          & 0                                  & 0                                                          &  {\color[HTML]{00629B} \textbf{15}} & {\color[HTML]{00629B} 6}          & 0                                  & 0                                  & 0                                                          & 0                                                          & 0                                 & 0                                  & 0                                  & 0                                  & {\color[HTML]{00629B} 1}           & 0                                  & 0                                  & 0                                  & 0                                 \\ \cline{2-32} 
\multicolumn{1}{|c|}{}                                   & \textbf{Ñ}    & 0                                                          & 0                                  & 0                                                          & 0                                                          & 0                                                          & 0                                  & 0                                 & 0                                                          & 0                                 & 0                                  & 0                                  & 0                                 & 0                                                          & 0                                  & 0                                                          & 0                                                          & {\color[HTML]{00629B} \textbf{9}} & 0                                  & 0                                  & 0                                                          & 0                                                          & 0                                 & 0                                  & 0                                  & 0                                  & 0                                  & 0                                  & 0                                  & 0                                  & 0                                 \\ \cline{2-32} 
\multicolumn{1}{|c|}{}                                   & \textbf{O}    & 0                                                          & 0                                  & 0                                                          & 0                                                          & 0                                                          & 0                                  & 0                                 & 0                                                          & 0                                 & 0                                  & 0                                  & 0                                 & 0                                                          & 0                                  & 0                                                          & 0                                                          & 0                                 & {\color[HTML]{00629B} \textbf{14}} & 0                                  & 0                                                          & 0                                                          & 0                                 & 0                                  & 0                                  & 0                                  & 0                                  & 0                                  & 0                                  & 0                                  & 0                                 \\ \cline{2-32} 
\multicolumn{1}{|c|}{}                                   & \textbf{P}    & 0                                                          & 0                                  & 0                                                          & 0                                                          & 0                                                          & 0                                  & 0                                 & 0                                                          & 0                                 & 0                                  & 0                                  & {\color[HTML]{00629B} 3}          & 0                                                          & 0                                  & 0                                                          & 0                                                          & 0                                 & 0                                  & {\color[HTML]{00629B} \textbf{11}} & 0                                                          & 0                                                          & 0                                 & 0                                  & 0                                  & 0                                  & 0                                  & {\color[HTML]{00629B} 4}           & 0                                  & 0                                  & 0                                 \\ \cline{2-32} 
\multicolumn{1}{|c|}{}                                   & \textbf{Q}    & 0                                                          & 0                                  & 0                                                          & 0                                                          & 0                                                          & 0                                  & 0                                 & 0                                                          & 0                                 & 0                                  & 0                                  & 0                                 & 0                                                          & 0                                  & 0                                                          & 0                                                          & 0                                 & 0                                  & 0                                  &  {\color[HTML]{00629B} \textbf{15}} & 0                                                          & 0                                 & 0                                  & 0                                  & 0                                  & 0                                  & 0                                  & 0                                  & 0                                  & 0                                 \\ \cline{2-32} 
\multicolumn{1}{|c|}{}                                   & \textbf{R}    & 0                                                          & 0                                  & 0                                                          & 0                                                          & 0                                                          & 0                                  & 0                                 & 0                                                          & 0                                 & 0                                  & 0                                  & 0                                 & 0                                                          & 0                                  & 0                                                          & 0                                                          & 0                                 & 0                                  & 0                                  & 0                                                          &  {\color[HTML]{00629B} \textbf{15}} & {\color[HTML]{00629B} 6}          & 0                                  & 0                                  & {\color[HTML]{00629B} 2}           & 0                                  & 0                                  & 0                                  & 0                                  & 0                                 \\ \cline{2-32} 
\multicolumn{1}{|c|}{}                                   & \textbf{RR}   & 0                                                          & 0                                  & 0                                                          & 0                                                          & 0                                                          & 0                                  & 0                                 & 0                                                          & 0                                 & 0                                  & 0                                  & 0                                 & 0                                                          & 0                                  & 0                                                          & 0                                                          & 0                                 & 0                                  & 0                                  & 0                                                          & 0                                                          & {\color[HTML]{00629B} \textbf{7}} & 0                                  & 0                                  & 0                                  & {\color[HTML]{00629B} 1}           & 0                                  & 0                                  & 0                                  & 0                                 \\ \cline{2-32} 
\multicolumn{1}{|c|}{}                                   & \textbf{S}    & 0                                                          & 0                                  & 0                                                          & 0                                                          & 0                                                          & {\color[HTML]{00629B} 1}           & 0                                 & 0                                                          & 0                                 & 0                                  & 0                                  & 0                                 & 0                                                          & 0                                  & 0                                                          & 0                                                          & 0                                 & {\color[HTML]{00629B} 1}           & 0                                  & 0                                                          & 0                                                          & 0                                 & {\color[HTML]{00629B} \textbf{14}} & 0                                  & 0                                  & 0                                  & 0                                  & 0                                  & 0                                  & 0                                 \\ \cline{2-32} 
\multicolumn{1}{|c|}{}                                   & \textbf{T}    & 0                                                          & 0                                  & 0                                                          & 0                                                          & 0                                                          & 0                                  & {\color[HTML]{00629B} 9}          & 0                                                          & 0                                 & 0                                  & 0                                  & 0                                 & 0                                                          & 0                                  & 0                                                          & 0                                                          & 0                                 & 0                                  & 0                                  & 0                                                          & 0                                                          & 0                                 & 0                                  & {\color[HTML]{00629B} \textbf{13}} & 0                                  & 0                                  & 0                                  & {\color[HTML]{00629B} 1}           & 0                                  & 0                                 \\ \cline{2-32} 
\multicolumn{1}{|c|}{}                                   & \textbf{U}    & 0                                                          & 0                                  & 0                                                          & 0                                                          & 0                                                          & 0                                  & 0                                 & 0                                                          & 0                                 & 0                                  & 0                                  & 0                                 & 0                                                          & 0                                  & 0                                                          & 0                                                          & 0                                 & 0                                  & 0                                  & 0                                                          & 0                                                          & 0                                 & 0                                  & 0                                  & {\color[HTML]{00629B} \textbf{11}} & {\color[HTML]{00629B} 2}           & 0                                  & 0                                  & 0                                  & 0                                 \\ \cline{2-32} 
\multicolumn{1}{|c|}{}                                   & \textbf{V}    & 0                                                          & 0                                  & 0                                                          & 0                                                          & 0                                                          & 0                                  & 0                                 & 0                                                          & 0                                 & 0                                  & 0                                  & 0                                 & 0                                                          & 0                                  & 0                                                          & 0                                                          & 0                                 & 0                                  & 0                                  & 0                                                          & 0                                                          & 0                                 & 0                                  & 0                                  & {\color[HTML]{00629B} 1}           & {\color[HTML]{00629B} \textbf{11}} & 0                                  & 0                                  & 0                                  & 0                                 \\ \cline{2-32} 
\multicolumn{1}{|c|}{}                                   & \textbf{W}    & 0                                                          & 0                                  & 0                                                          & 0                                                          & 0                                                          & 0                                  & 0                                 & 0                                                          & 0                                 & 0                                  & 0                                  & 0                                 & 0                                                          & 0                                  & 0                                                          & 0                                                          & 0                                 & 0                                  & {\color[HTML]{00629B} 3}           & 0                                                          & 0                                                          & 0                                 & 0                                  & 0                                  & 0                                  & 0                                  & {\color[HTML]{00629B} \textbf{10}} & 0                                  & 0                                  & 0                                 \\ \cline{2-32} 
\multicolumn{1}{|c|}{}                                   & \textbf{X}    & 0                                                          & 0                                  & 0                                                          & 0                                                          & 0                                                          & 0                                  & 0                                 & 0                                                          & 0                                 & 0                                  & 0                                  & 0                                 & 0                                                          & 0                                  & 0                                                          & 0                                                          & 0                                 & 0                                  & 0                                  & 0                                                          & 0                                                          & 0                                 & 0                                  & 0                                  & 0                                  & 0                                  & 0                                  & {\color[HTML]{00629B} \textbf{13}} & 0                                  & 0                                 \\ \cline{2-32} 
\multicolumn{1}{|c|}{}                                   & \textbf{Y}    & 0                                                          & 0                                  & 0                                                          & 0                                                          & 0                                                          & 0                                  & 0                                 & 0                                                          & 0                                 & {\color[HTML]{00629B} 1}           & 0                                  & 0                                 & 0                                                          & 0                                  & 0                                                          & 0                                                          & 0                                 & 0                                  & 0                                  & 0                                                          & 0                                                          & 0                                 & 0                                  & 0                                  & 0                                  & 0                                  & 0                                  & 0                                  & {\color[HTML]{00629B} \textbf{13}} & 0                                 \\ \cline{2-32} 
\multicolumn{1}{|c|}{}                                   & \textbf{Z}    & 0                                                          & 0                                  & 0                                                          & 0                                                          & 0                                                          & 0                                  & 0                                 & 0                                                          & 0                                 & 0                                  & {\color[HTML]{00629B} 2}           & 0                                 & 0                                                          & 0                                  & 0                                                          & 0                                                          & 0                                 & 0                                  & 0                                  & 0                                                          & 0                                                          & 0                                 & 0                                  & 0                                  & {\color[HTML]{00629B} 1}           & 0                                  & 0                                  & 0                                  & 0                                  & {\color[HTML]{00629B} \textbf{8}} \\ \cline{2-32} 
\multicolumn{1}{|c|}{\multirow{-31}{*}{\rotatebox[origin=c]{90}{Result Obtained}}} & \textbf{Other} & {\color[HTML]{00629B} 2}                                   & {\color[HTML]{00629B} 1}           & {\color[HTML]{00629B} 1}                                   & {\color[HTML]{00629B} 1}                                   & {\color[HTML]{00629B} 1}                                   & {\color[HTML]{00629B} 1}           & {\color[HTML]{00629B} 1}          & {\color[HTML]{00629B} 1}                                   & {\color[HTML]{00629B} 1}          & {\color[HTML]{00629B} 1}           & {\color[HTML]{00629B} 1}           & {\color[HTML]{00629B} 1}          & {\color[HTML]{00629B} 1}                                   & {\color[HTML]{00629B} 1}           & {\color[HTML]{00629B} 1}                                   & {\color[HTML]{00629B} 1}                                   & {\color[HTML]{00629B} 1}          & {\color[HTML]{00629B} 1}           & {\color[HTML]{00629B} 1}           & {\color[HTML]{00629B} 1}                                   & {\color[HTML]{00629B} 1}                                   & {\color[HTML]{00629B} 1}          & {\color[HTML]{00629B} 1}           & {\color[HTML]{00629B} 1}           & {\color[HTML]{00629B} 1}           & {\color[HTML]{00629B} 1}           & {\color[HTML]{00629B} 1}           & {\color[HTML]{00629B} 1}           & {\color[HTML]{00629B} 1}           & {\color[HTML]{00629B} 1}          \\ 
\noalign{\hrule height 1.0pt}
\end{tabular}}}
\end{table}

\subsubsection{Individual Letter Error Analysis}

\begin{figure}[h!]
  \centering
    \includegraphics[width=0.6\textwidth]{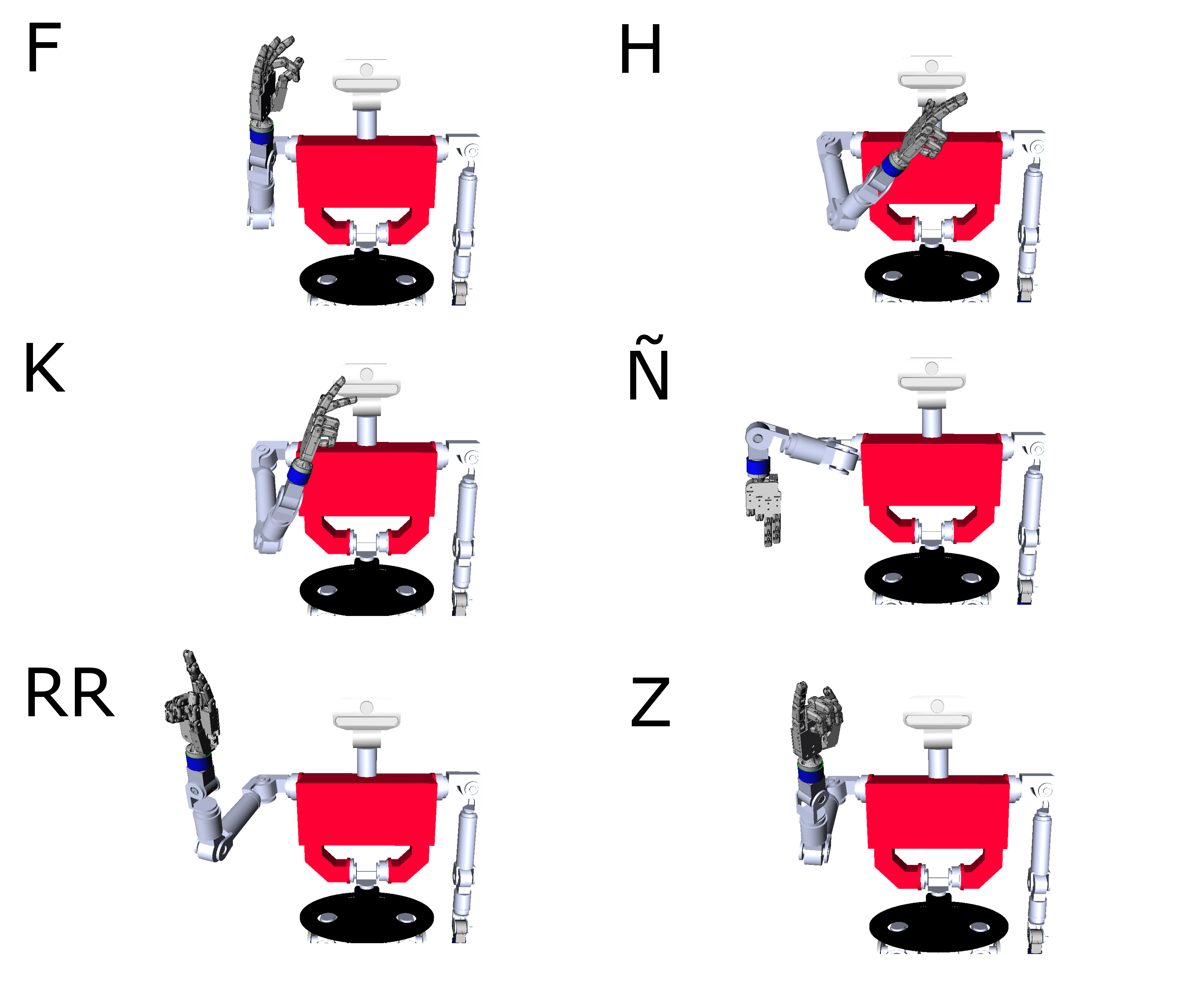}
  \caption{Initial frames of challenging letters in dactylology. \textit{Dactylology obtained from CNSE Foundation for the Suppression of Communication Barriers images and signs LSE database}.} 
  \label{conflictive}
\end{figure}

Taking a deeper look at the matrix helps to clarify the source of errors in individual letter recognition. The most controversial letters, of which the initial frames are shown in Figure \ref{conflictive}, are F, H, K, Ñ, RR and Z, with a correct answer rate less or equal to 75\%. An independent study for each letter is convenient to identify causes of confusion.

\begin{itemize}

    \item {Letter F is the most erratic letter of the experiment. It was mistaken for letter T in 60\% of the attempts. These letters have relatively similar finger position configuration, as can be seen in Figure \ref{signs}. It is remarkable that only 13\% of users mistaken letter T for letter F. Reviewing the F simulation file its has been noticed that thumb position may have resulted confusing, and it has been modified for future experiments. }

    \item {Letter H is mistaken for letter CH in almost 27\% of the attempts. Both letters have the same finger position configuration, but they differ in arm movement. Letter CH has not been mistaken for letter H in any attempt so the arm movement for CH was pronounced after this analysis.}

    \item {Letter K is mistaken for letters H and P in 20\% of the attempts each. In this case, there is over a 7\% of coincidence regarding letter H and no coincidence at all regarding letter P. Letter K is a specially complex case, since the position of the middle finger is not so evident as it is in other letters and there must have been some implementation errors that should be rectified with the help of an LSE expert.}

    \item {Letter Ñ is mistaken for letter N 40\% of the attempts. The only difference between these letters is that letter N is static and letter Ñ requires movement. The solution provided to decrease this error is to make the movement more noticeable to avoid being confused with a letter transition.}

    \item {Letter RR is mistaken for letter R in 40\% of the attempts. The casuistic is exactly the same as in the Ñ-N case. Therefore, the same solution is provided.}

    \item {\textls[-5]{Letter Z is mistaken for letter J in 40\% of the attempts. Both letters need motion and they share finger position configuration with letter I. Letter J performs a circular movement while letter Z performs a zig-zag movement. The second one was developing this movement in an almost horizontal plane, so it was not easily understandable. The solution was to change the angle of movement execution.}}
    
\end{itemize}

Some other letters show small inaccuracies of which sources are not as immediately perceptible as these previous ones, so further analysis is required to find new root causes.

\subsubsection{Age Influence in Dactylology}

\begin{figure}[H]
  \centering
    \includegraphics[width=0.7\textwidth]{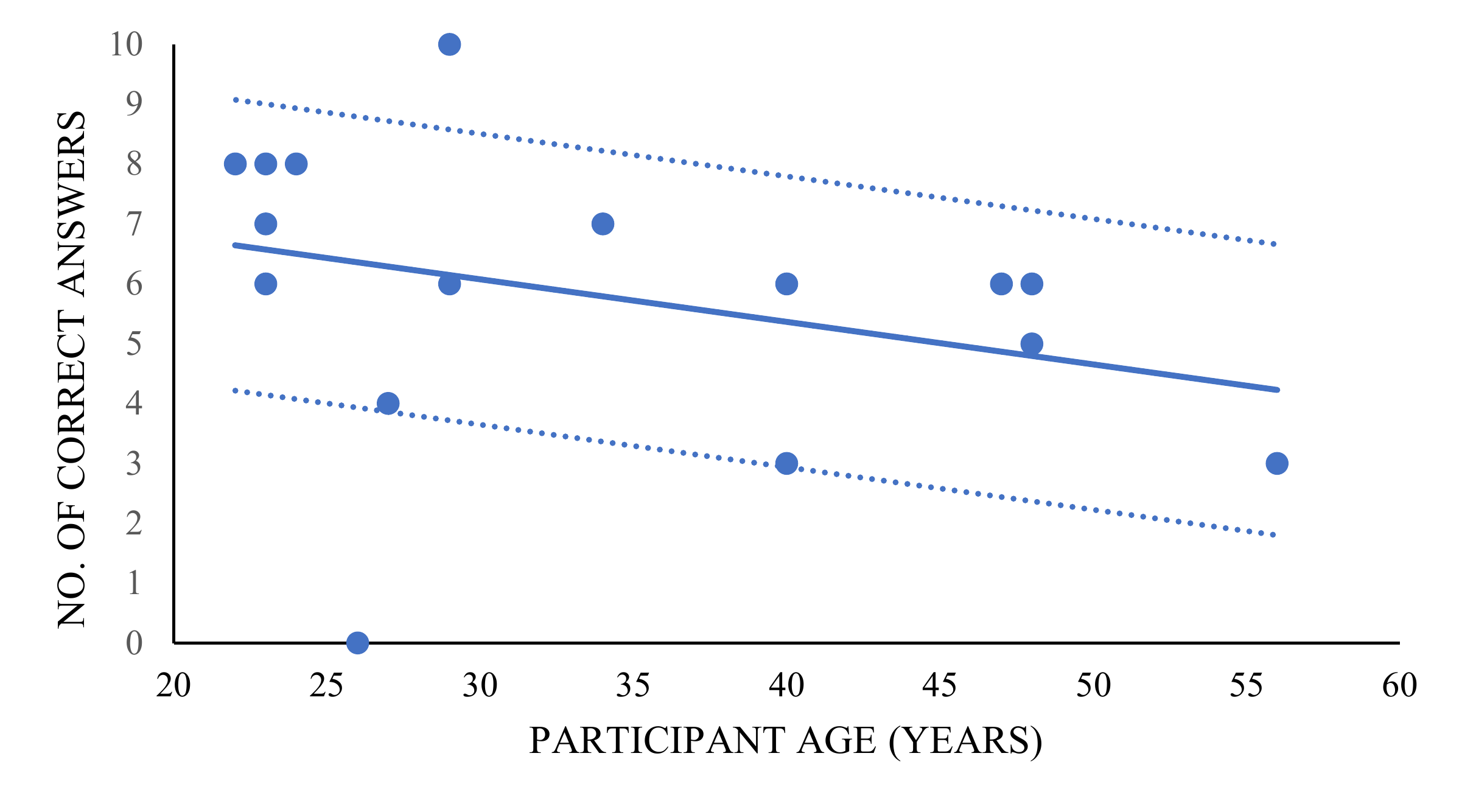}
  \caption{Graph of number of correct answers per participant in dactylology test versus participant's age (years) with linear trendline and linear regression channel which contains approximately 68\% of all answers.}
  \label{dactyl-age}
\end{figure}

Figure \ref{dactyl-age} shows the relation between the number of correct answers and the age of the users. The negative slope of the linear trendline shows a light tendency towards misunderstanding the dactylology developed by TEO in relation to age increase. To measure letter transition understanding, the answer is considered correct only if the user is able to understand the complete set of three letters, which means that the movements between letters have not influenced the correct perception of the dactylology.

The regression channel, which is the area between dotted lines in Figure \ref{dactyl-age}, is based on the linear regression that represents a simple trendline that is projected using the least squares method. Consequently, this line turns out to be an average line of the correct answer rate that is changing. It can be considered as an ``equilibrium'' result line, while any deviation from it up or down indicates the higher activity of correct or wrong answers, respectively \cite{b18}. The distance between the channel bands and the regression line is equal to the standard deviation value of the correct answer rate with respect to the regression line. The upper and lower channel lines therefore contain between themselves approximately 68\% of all user answer data.

\newpage

For this dactylology test $d$, the trendline equation and the coefficient of determination $R^{2}_d$ obtained by the least squares method are shown in Equations (\ref{eq1}) and  (\ref{eq2}), respectively. Since the regression line is relatively far from some of the points, the $R^{2}_d$ of the regression is quite low.
\begin{equation}
    y_d = -0.071x_d+ 8.2056
    \label{eq1}
\end{equation}
\begin{equation}
{\displaystyle R^{2}_d={\sigma _{X_dY_d}^{2} \over \sigma _{X_d}^{2}\sigma _{Y_d}^{2}}} = 0.1076
\label{eq2}
\end{equation}

The standard deviation $s_d$ for the regression channel included in Figure \ref{dactyl-age} is shown in Equation~(\ref{eq3}). Using this value and Equation (\ref{eq1}), upper and lower lines on the regression channel are drawn.
\begin{equation}
{\displaystyle s_d={\sqrt {\frac {\sum _{i=1}^{N}(y_{d_i}-{\overline {y_d}})^{2}}{N-1}}}=2.4281}
\label{eq3}
\end{equation}

For these sets of three words, this standard deviation shows that a range of dactylology understanding approximately between 42\% and 90\% can be expected from users in their early twenties, in comparison to the 18--67\% approximated range for middle aged people.

It should be taken into consideration for this analysis that there is an outlier due to one user which, through a manual review of the answers, can be determined to have answered the comprehension tests arbitrarily. Consequently, if this outlying data is omitted, the data presented in Equations (\ref{eq1})--(\ref{eq3}) presents the variations presented in Equations (\ref{eq1a})--(\ref{eq3a}), respectively.
\begin{equation}
    y_{d'} = -0.0996x_{d'}+ 9.608
    \label{eq1a}
\end{equation}
\begin{equation}
{\displaystyle R^{2}_{d'}= 0.3453}
\label{eq2a}
\end{equation}
\begin{equation}
{\displaystyle s_{d'}=1.9346
}
\label{eq3a}
\end{equation}

As expected, this change presents a steeper negative slope and the coefficient of determination $R^2_d$ has increased more than three times while the standard deviation $s_d$ has decreased. As this model is more adjusted to the variable when the outlier is omitted, it can be concluded that the tendency to misunderstand the dactylology in relation to age is more pronounced than previously stated. 

\subsection{Basic House Vocabulary}
Vocabulary test results were significantly more positive than the dactylology ones. The correct answer rate can be checked in the confusion matrix shown in Table \ref{vocab}.

The average of correct answers per user is 13.3, which means a 83\% of success rate (133 correct answers over 160 answers). The lowest understood word achieved a 62.5\% of correct answers (10 correct answers over 16 answers), so almost two thirds of the users were able to understand even the most challenging words. This result in vocabulary understanding was expected, since word signing, in this particular case, does not require a high level of detail and it is more figurative than fingerspelling. 

\begin{table}[H]
\centering
\small
\caption{Confusion matrix: basic house vocabulary. The elements of diagonal, which represent correct answers, are marked in bold. Elements with a shaded background mean 100\% correct answers (15 correct, and 16 in the exceptional case where there was a correct answer within the outlying data).}
\label{vocab}
\setlength{\tabcolsep}{3pt}
\resizebox{\textwidth}{!}
{\begin{tabular}{|c|c|c|c|c|c|c|c|c|c|c|c|}
\noalign{\hrule height 1.0pt}

                                                         & \textbf{}           & \multicolumn{10}{c|}{\textbf{Result Expected}}                                                                                                                                                                                                                                                                                                                                                                                   \\ \cline{3-12} 
\textbf{}                                                &                     & \textbf{Table}                         & \textbf{Door}                          & \textbf{Bedroom}                       & \textbf{Closet}                        & \textbf{Telephone}                     & \textbf{Machine}                       & \textbf{Kitchen}                       & \textbf{Clothes}                       & \textbf{Iron}                          & \textbf{\begin{tabular}[c]{@{}c@{}}Living room \end{tabular}}                   \\ \cline{2-12} 
\multicolumn{1}{|c|}{}                                   & \textbf{Table}       & {\color[HTML]{00629B} \textbf{14}} & 0                                 & 0                                 & 0                                 & 0                                 & 0                                & 0                                & {\color[HTML]{00629B} 1}          & 0                                & 0                                 \\ \cline{2-12} 
\multicolumn{1}{|c|}{}                                   & \textbf{Door}     & 0                                & {\color[HTML]{00629B} \textbf{10}} & {\color[HTML]{00629B} 2}          & {\color[HTML]{00629B} 2}          & 0                                 & 0                                 & {\color[HTML]{00629B} 1}          & 0                                & 0                                & 0                                 \\ \cline{2-12} 
\multicolumn{1}{|c|}{}                                   & \textbf{Bedroom} & 0                                & {\color[HTML]{00629B} 2}          & {\color[HTML]{00629B} \textbf{14}} & 0                                & 0                                 & 0                                 & 0                                 & 0                                 & 0                               & 0                                 \\ \cline{2-12} 
\multicolumn{1}{|c|}{}                                   & \textbf{Closet}    & 0                                & {\color[HTML]{00629B} 4}          & 0                                 & {\color[HTML]{00629B} \textbf{11}} & 0                                & 0                              & 0                                 & 0                                 & 0                                & {\color[HTML]{00629B} 1}          \\\cline{2-12} 
\multicolumn{1}{|c|}{}                                   & \textbf{Telephone}   & 0                                 & 0                                 & 0                                & 0                                 & {\color[HTML]{00629B}  \textbf{16}} & 0                                 & 0                                 & 0                                 & 0                                 & {\color[HTML]{00629B} 1}          \\ \cline{2-12} 
\multicolumn{1}{|c|}{}                                   & \textbf{Machine}    & 0                                 & 0                                 & 0                                 & 0                                 & 0                                 & {\color[HTML]{00629B} \textbf{12}} & 0                                 & 0                                 & {\color[HTML]{00629B} 1}          & {\color[HTML]{00629B} 2}          \\ \cline{2-12} 
\multicolumn{1}{|c|}{}                                   & \textbf{Kitchen}     & {\color[HTML]{00629B} 1}          & 0                                 & 0                                 & 0                                 & 0                                 & 0                                 & {\color[HTML]{00629B} \textbf{14}} & 0                                 & 0                                 & 0                                 \\  \cline{2-12}
\multicolumn{1}{|c|}{}                                   & \textbf{Clothes}       & 0                                 & 0                                 & 0                                 & 0                                 & 0                                 & {\color[HTML]{00629B} 3}          & 0                                 & { \color[HTML]{00629B} \textbf{15}} & 0                                 & 0                                 \\  \cline{2-12}  
\multicolumn{1}{|c|}{}                                   & \textbf{Iron}    & 0                                 & 0                                 & 0                                 & 0                                 & 0                                 & 0                                 & {\color[HTML]{00629B} 1}          & 0                                 & { \color[HTML]{00629B} \textbf{15}} & 0                                 \\ \cline{2-12} 
\multicolumn{1}{|c|}{\multirow{-10}{*}{\rotatebox[origin=c]{90}{Result Obtained}}} & \textbf{\begin{tabular}[c]{@{}c@{}}Living room \end{tabular}}      & {\color[HTML]{00629B} 1}          & 0                                 & 0                                 & {\color[HTML]{00629B} 3}          & 0                                 & {\color[HTML]{00629B} 1}          & 0                                 & 0                                 & 0                                 & {\color[HTML]{00629B} \textbf{12}} \\
 \noalign{\hrule height 1.0pt}
\end{tabular}}
\end{table}

\subsubsection{Vocabulary Error Analysis}

In order to detect some irregularities and check if the groups of similar words produced confusion among users, a detailed error analysis is developed.

\begin{itemize}
\item {``Machine'' and  ``clothes'' are two words in LSE that are similar, since the main difference between them is the position of the hand, but the arms develop relatively the same movement. ``Machine'' was mistaken for ``clothes'' in almost 19\% of the answers. The word ``clothes'' was however never mistaken for ``machine''. Since the difference between both words is a matter of open/close fist variation it is possible that users are not so used to the word ``machine'' or even that the first word that appeared in the drop-down list was ``clothes''.}
\item {``Door'',  ``kichen''  and  ``closet'' are words that require a similar arm and hand movement, with some variations in the order the hands are positioned. ``Door'' was mistaken for ``closet'' 25\% of the attempts and only a 13\% in the inverse order. This difference may be attributed to simulation, since the hand position order was correctly developed. ``Kitchen'' was mistaken for ``door'' over a 6\% of the attempts but not a single time in the opposite way, so it is not considered significant. There were no connections at all between ``closet'' and ``kitchen'' in either direction. }
\item {``Bedroom'' and  ``table'' could have been confused since they share similar movements, but they were not confused at any time.}
\item {``Living  room''  and  ``telephone'' are two words that require signing in the head area. ``Telephone'' was identified 100\% of the attempts, which is an interesting rate, considering there is one user who submitted most of their answers wrong. The reason for this accuracy may be explained through the fact that the Spanish sign for ``telephone'' could be understood internationally without any LSE knowledge. ``Living room'' was not confused at any time with ``telephone'', but it was confused approximately 13\% of the attempts with ``machine'', which is an outcome that cannot be explained from the consulted LSE signing database point of view.} 

\end{itemize}

An unexpected result was the confusion between ``closet'' and ``living room'', with about 19\% error rate in the mentioned order, and just a 6\% in the inverse order. There is no relation at all between the way both words were simulated, so it may indicate an implementation error or may be biased by frequency of everyday use. 

The only independent word, which is ``iron'', did not represent any challenge for users, since it presented a 94\% of success rate. Some other words also shown small inaccuracies, which sources are relatively difficult to determine.

\subsubsection{Age Influence in Vocabulary}

\begin{figure}[H]
  \centering
    \includegraphics[width=0.7\textwidth]{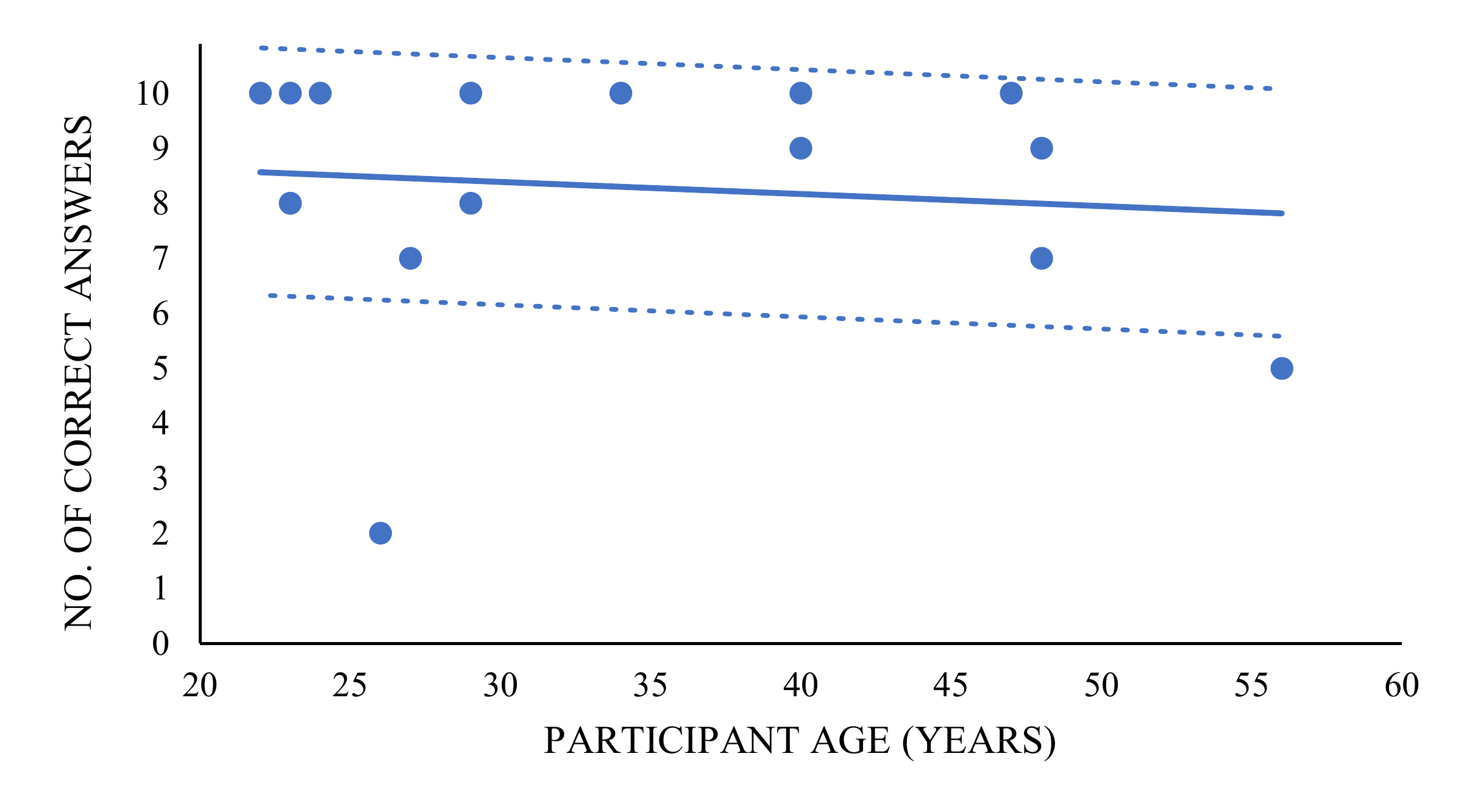}
  \caption{Graph of number of correct answers per participant in vocabulary test versus participant's age (years) with linear trendline and linear regression channel which contains approximately 68\% of all answers.}
  \label{vocab-age}
\end{figure}

Figure \ref{vocab-age} shows the relation between the rate of correct answers and the age of the users. The~negative trendline slope shows a even lighter tendency to misunderstand the vocabulary in relation to age than the one presented in Figure \ref{dactyl-age}.

For this vocabulary test $v$, the resulting equation used to draw the trendline and the coefficient of determination $R^{2}_v$ are shown in Equations (\ref{eq4}) and  (\ref{eq5}). As expected, the slope of the negative trendline is less than one third of the dactylology trendline slope, which means that the tendency to misunderstand sign language using words in relation to age is almost insignificant. Since the regression line is relatively far from a high percentage of the points, the $R^{2}_v$ of the regression is quite~low.
\begin{equation}
   y_{v} = -0.022x_{v} + 9.0529 
   \label{eq4}
\end{equation}
\begin{equation}
   {\displaystyle R^{2}_{v}={\sigma _{X_{v}Y_{v}}^{2} \over \sigma _{X_{v}}^{2}\sigma _{Y_{v}}^{2}}} = 0.0121 
   \label{eq5}
\end{equation}

The standard deviation for the regression channel included in Figure \ref{vocab-age} is shown in Equation (\ref{eq6}). Using this value and Equation (\ref{eq4}), upper and lower lines on the regression channel can be drawn.
\begin{equation}
{\displaystyle s_{v}={\sqrt {\frac {\sum _{i=1}^{N}(y_{v_i}-{\overline {y_v}})^{2}}{N-1}}}=2.2425}
\label{eq6}
\end{equation}

This standard deviation shows that a range of vocabulary understanding between 63\% and 100\% can be expected from users in their twenties, quite similar to the 56--100\% range for middle aged~people.

If the outlier is also not being considered in this case, the data presented in Equations (\ref{eq4})--(\ref{eq6}) presents some variations, which are presented in Equations (\ref{eq4a})--(\ref{eq6a}), respectively.
\begin{equation}
    y_{v'} = -0.0511x_{v'}+ 10.482
    \label{eq4a}
\end{equation}
\begin{equation}
{\displaystyle R^{2}_{v'}= 0.1447}
\label{eq5a}
\end{equation}
\begin{equation}
{\displaystyle s_{v'}=1.533747356
}
\label{eq6a}
\end{equation}

As occurred with dactylology, this change presents over a double steeper negative slope, and the coefficient of determination $R^2_v$ has increased almost twelve times, while the standard deviation $s$ has decreased significantly. As this model is more adjusted to the variable when the outlier is omitted, it can be concluded that the tendency to misunderstand the dactylology in relation to age is slightly more pronounced than previously stated. 

Considering the obtained data, it can be concluded that this high vocabulary understanding correct answer rate and this small pronounced slope in comparison with the dactylology outcomes may be due, not only to the signing simplicity, but also to the fact that letters were displayed in sets of three, while words were tested independently and not in a sentence in order to simplify analysis, so it is understandable that the error rate decreases. 

\subsection{User Satisfaction}

Table \ref{satisftable} shows the satisfaction questionnaire individual results, sorted by age and measured in a [$-$2, 2] Likert scale. Average overall user satisfaction over this experimental work results in a promising 0.78 (69.5\%), roughly between a neutral and positive position.

\begin{figure}[H]
  \centering
    \includegraphics[width=0.75\textwidth]{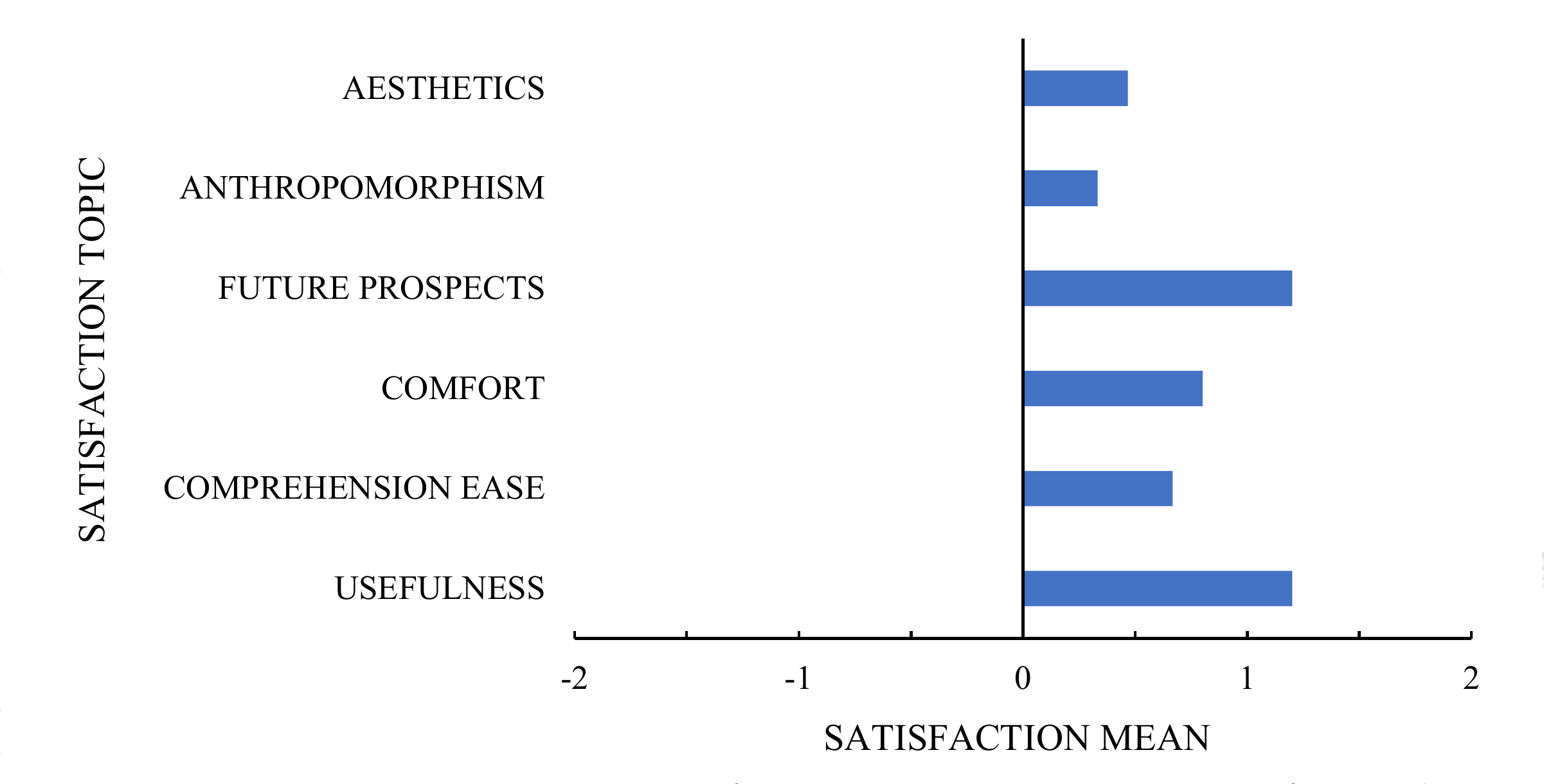}
  \caption{Overall satisfaction classified by topic. A five level Likert scale has been used, where the values mean: ($-$2) Strongly disagree, ($-$1) Disagree, (0) Neither agree nor disagree, (1) Agree, (2) Strongly agree.}
  \label{overallsatisfaction}
\end{figure}

This data is grouped and analysed with the purpose of drawing relevant conclusions. Figure~\ref{overallsatisfaction} gives a breakdown of this outcome, where no negative mean values can be observed, but some relevant different satisfaction levels are found.

\begin{table}[h!]
\centering
\small
\caption{Satisfaction topics outcome classified by age in ascending order. A five level Likert scale has been used, where the values mean: ($-$2) Strongly disagree, ($-$1) Disagree, (0) Neither agree nor disagree, (1) Agree, (2) Strongly agree.}
\label{satisftable}
\setlength{\tabcolsep}{3pt}
\resizebox{\textwidth}{!}{\begin{tabular}{|c|c|c|c|c|c|c|c|c|}
\noalign{\hrule height 1.0pt}

                                             &                                  & \multicolumn{6}{c|}{\textbf{Satisfaction topics}}                                                                                                           &                                                                                                  \\ \cline{3-8}
                                             & {\color[HTML]{000000} \textbf{}} & \textbf{Usefulness} & \textbf{\begin{tabular}[c]{@{}c@{}}Comprehension ease \end{tabular}} & \textbf{Comfort} & \textbf{\begin{tabular}[c]{@{}c@{}}Future prospects\end{tabular}} & \textbf{Anthropomorphism} & \textbf{Aesthetics} & \multirow{-2}{*}{\textbf{\begin{tabular}[c]{@{}c@{}}Satisfaction mean \end{tabular}}} \\
                                        \cline{2-9}
\multicolumn{1}{|c|}{}                       & \textbf{22}                      & \phantom{-}1                   & \phantom{-}2                           & \phantom{-}0                & \phantom{-}1                         & \phantom{-}1                         & \phantom{-}1                   & \phantom{-}1.00                                                                                             \\ \cline{2-9} 
\multicolumn{1}{|c|}{}                       & \textbf{23}                      & \phantom{-}2                   & \phantom{-}2                           & \phantom{-}2                & \phantom{-}2                         & \phantom{-}2                         & \phantom{-}2                   & \phantom{-}2.00                                                                                             \\ \cline{2-9} 
\multicolumn{1}{|c|}{}                       & \textbf{23}                      & \phantom{-}2                   & \phantom{-}2                           & \phantom{-}2                & \phantom{-}2                         & \phantom{-}2                         & \phantom{-}2                   & \phantom{-}2.00                                                                                             \\ \cline{2-9} 
\multicolumn{1}{|c|}{}                       & \textbf{23}                      & \phantom{-}1                   & -1                          & \phantom{-}1                & \phantom{-}2                         & -1                        & -1                  & \phantom{-}0.17                                                                                             \\ \cline{2-9} 
\multicolumn{1}{|c|}{}                       & \textbf{24}                      & \phantom{-}2                   & \phantom{-}1                           & \phantom{-}1                & \phantom{-}2                         & \phantom{-}1                         & \phantom{-}0                   & \phantom{-}1.00                                                                                             \\ \cline{2-9} 
\multicolumn{1}{|c|}{}                       & \textbf{26}                      & \phantom{-}1                   & \phantom{-}1                           & \phantom{-}1                & \phantom{-}1                         & \phantom{-}1                         & \phantom{-}1                   & \phantom{-}1.00                                                                                             \\ \cline{2-9} 
\multicolumn{1}{|c|}{}                       & \textbf{27}                      & \phantom{-}1                   & \phantom{-}0                           & \phantom{-}2                & \phantom{-}2                         & \phantom{-}1                         & \phantom{-}2                   & \phantom{-}1.33                                                                                             \\ \cline{2-9} 
\multicolumn{1}{|c|}{}                       & \textbf{29}                      & \phantom{-}2                   & \phantom{-}2                           & \phantom{-}2                & \phantom{-}2                         & \phantom{-}2                         & \phantom{-}2                   & \phantom{-}2.00                                                                                             \\ \cline{2-9} 
\multicolumn{1}{|c|}{}                       & \textbf{29}                      & \phantom{-}2                   & \phantom{-}2                           & \phantom{-}0                & \phantom{-}1                         & \phantom{-}1                         & -1                  & \phantom{-}0.83                                                                                             \\ \cline{2-9} 
\multicolumn{1}{|c|}{}                       & \textbf{34}                      & \phantom{-}2                   & \phantom{-}2                           & \phantom{-}2                & \phantom{-}2                         & \phantom{-}1                         & \phantom{-}2                   & \phantom{-}1.83                                                                                             \\ \cline{2-9} 
\multicolumn{1}{|c|}{}                       & \textbf{40}                      & \phantom{-}1                   & -1                          & -1               & \phantom{-}1                         & -1                        & -1                  & -0.33                                                                                            \\ \cline{2-9} 
\multicolumn{1}{|c|}{}                       & \textbf{40}                      & \phantom{-}0                   & -1                          & -1               & -1                        & -1                        & \phantom{-}0                   & -0.67                                                                                            \\ \cline{2-9} 
\multicolumn{1}{|c|}{}                       & \textbf{47}                      & \phantom{-}1                   & \phantom{-}1                           & \phantom{-}1                & \phantom{-}1                         & \phantom{-}1                         & \phantom{-}1                   & \phantom{-}1.00                                                                                             \\ \cline{2-9} 
\multicolumn{1}{|c|}{}                       & \textbf{48}                      & \phantom{-}2                   & -1                          & \phantom{-}2                & \phantom{-}2                         & \phantom{-}0                         & -1                  & \phantom{-}0.67                                                                                             \\ \cline{2-9} 
\multicolumn{1}{|c|}{}                       & \textbf{48}                      & \phantom{-}1                   & \phantom{-}1                           & -1               & \phantom{-}0                         & -1                        & -1                  & -0.17                                                                                            \\ \cline{2-9} 
\multicolumn{1}{|c|}{\multirow{-16}{*}{{\rotatebox[origin=c]{90}{Age}}}} & \textbf{56}                      & -1                  & \phantom{-}0                           & -1               & -1                        & -2                        & -2                  & -1.17                                                                                          \\ 
\noalign{\hrule height 1.0pt}
\end{tabular}}
\end{table}

Top valued topics were future prospects and usefulness, with a 1.2 average satisfaction or, which is the same, an 80\% positive feedback. This result demonstrates the user willingness to use this technology, and the high level of expectation the use of LSE with a humanoid robot this first contact has generated.  

Comfort and comprehension ease, with 0.8 (70\%) and 0.7 (67\%) of average satisfaction, respectively, occupy the following positions in the ranking. A reasonable explanation to find these topics lower rated than the previous ones is that there are various letters and vocabulary which have presented some understanding difficulties and have led to confusion. In any case, as proved, these minor inconveniences have not influenced the user expectation. Finally, the least favourable marks are aesthetics and anthropomorphism, with a 62\% and 58\%. These topics are closely associated to the robot appearance. Since TEO is still being developed at both software and hardware level, it is comprehensible that there are divergent opinions about the way it looks. In either case, this nearly neutral anthropomorphism user perception should not be interpreted as a negative outcome, since resemblance to a human being is not only unnecessary, but also a characteristic to be avoided in assistive robotics.

\subsubsection{Age Influence in Satisfaction}

\begin{figure}[H]
  \centering
    \includegraphics[width=0.8\textwidth]{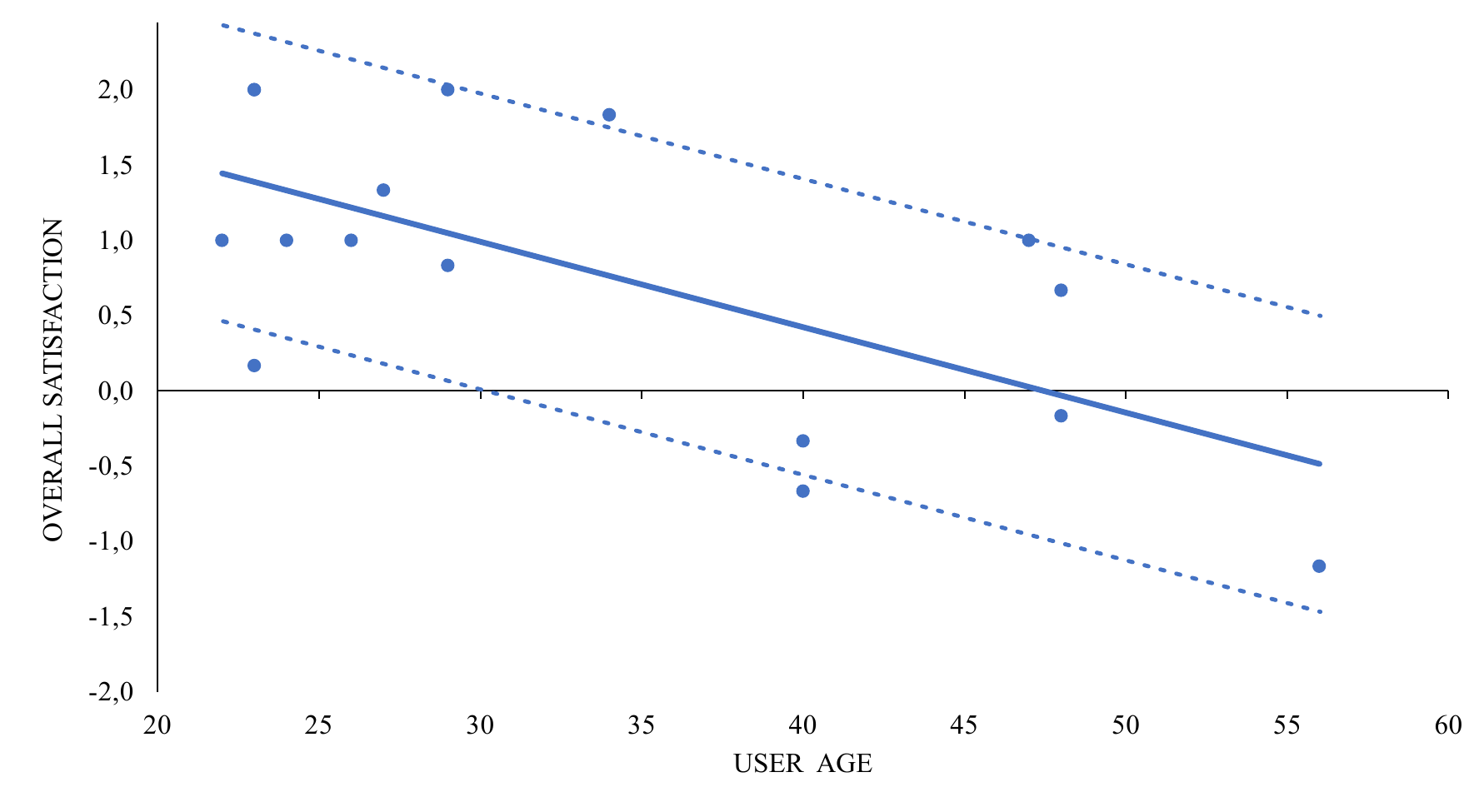}
  \caption{Graph of user satisfaction relation versus age (years) with linear trendline and linear regression channel which contains approximately 68\% of all answers. A five level Likert scale has been used, where the values mean: ($-$2) Strongly disagree, ($-$1) Disagree, (0) Neither agree nor disagree, (1) Agree, (2) Strongly agree.}
  \label{satisfaction}
\end{figure}

Age influence in overall user satisfaction is related to its influence in dactylology and vocabulary understanding. Figure \ref{satisfaction}, where the satisfaction-age relation is shown, presents a negative trendline that goes through the neutral line, so it is the first graphic in which the trend drops almost a 50\% from the youngest to the oldest age.

\newpage

For this satisfaction questionnaire $s$, the resulting equation used to draw the trendline and the $R^{2}_s$ value are shown in Equation (\ref{eq7}) and Equation (\ref{eq8}).
\begin{equation}
   y_{s} = -0.0568x_{s} + 2.6947
   \label{eq7}
\end{equation}
\begin{equation}
   {\displaystyle R^{2}_{s}={\sigma _{X_{s}Y_{s}}^{2} \over \sigma _{X_{s}}^{2}\sigma _{Y_{s}}^{2}}} = 0.42 
   \label{eq8}
\end{equation}

The standard deviation $s_s$ for the regression channel included in Figure \ref{satisfaction} is shown in Equation~(\ref{eq9}). Using this value and Equation (\ref{eq7}), the upper and lower lines on the regression channel can be drawn.
\begin{equation}
{\displaystyle s_{s}={\sqrt {\frac {\sum _{i=1}^{N}(y_{s_i}-{\overline {y_s}})^{2}}{N-1}}}=0.9827}
 \label{eq9}
\end{equation}

This standard deviation shows that between 62\% and 100\% user satisfaction can be expected from users in their twenties, in comparison to a 27\% to 62\% range for middle aged people. The input from the user that previously provided outlying data is not affecting these results dramatically, and is therefore taken as a valid part of the experiment.

Taking a deeper look into Table \ref{satisftable}, further conclusions can be drawn. For users between 20 and 40 years old, the main disadvantages of this work are related to anthropomorphism and aesthetics; while users between 40 and 60 years old, also find quite inconvenient the comprehension ease and comfort. This satisfaction distribution helps to identify  what fields need to be improved to reach all users. Although it is reasonable to expect that people around this second group may not be so used to technology as the millennial generation, and that their answers could also be influenced by the online test format, it is convenient to reach a universal level of ease of understanding for the wide range of people that could need to communicate with the robot.

\subsection{User Additional Comments}

Approximately 63\% of users left additional optional comments with their personal opinions. This feedback is highly valuable, since it offers the opportunity to detect further aspects that need improvement, while it sets the basis for communication error understanding.

\subsubsection{Alternatives for Human--Robot Interaction}

End-users are asked to provide alternative possibilities to the two options considered in this paper: subtitles and sign language. This question was answered by 31\% of the participants, of which only one user actually provided an alternative solution: using a robotic mouth which is able to enunciate accurately while using sign language. The rest of opinions encouraged the use of sign language as the best available option.

\subsubsection{Justification of Preference}

User preference justification must be divided between users which finally selected sign language and the one that selected subtitles as the ideal option after TEO signing demonstration.

The reasons for selecting sign language can be summed up in the following points:

\begin{itemize}
    \item Sign language is clearer.
    \item Sign language is more understandable.
    \item Many deaf people find it difficult to read or interpret a text.
    \item Interpreting signing is effortless for the user, since they use sign language in a daily basis.
\end{itemize}

Two points were shared by one user to justify the subtitles selection:

\begin{itemize}
    \item Lack of facial expression and lip-speaking.
    \item Depending on the context, a sign can have several different meanings.
\end{itemize}

Overall, there are various reasons regarding comfort which lead people to prefer interacting with the robot via sign language. Nevertheless, it must be taken into consideration that there are some drawbacks, such as the lack of facial expression, which may hinder the communication.

\subsubsection{Proposals for Improvement}

The submitted suggestions were of major importance to detect which areas required improvement. There were several individual comments which stressed that the representation of LSE was clear and understanding the robot is a matter of practice. The remaining comments are listed bellow.

\begin{itemize}
    \item Some words are not sign language or they are not used anymore, such as ``living room''. This note highlights the importance of working with people specialised in LSE to implement this language.
    \item Human-like appearance is demanded. Human misconception of what to expect from a robot may be biased due to science fiction culture and may lead some users to feel disillusioned by the humanoid appearance or ``behaviour''.
    \item Hand motion seems too rigid. The robot is made of hard materials and actuated by electrical actuators, so it is complicated to reproduce a smooth motion as human muscles can perform. 
    \item Bigger size of the images in the form would be required. This is an important point since it could justify the increasing failure tendency in understanding the robot related to age, considering that the decline of vision generally associated with age. 
\end{itemize}

\section{Discussion}

Regarding the developed research, detected several challenges that may be addressed in order to enhance the analysis of the results can be described. As mentioned, only the user's age was considered as the main characteristic to evaluate the tests and questionnaire outcome. However, it would be highly useful to ask the participants for their education level and to let them rate their frequency of use and familiarity with technology. These elements could help factor out some outlying responses and classify the data more accurately, especially in future studies where a larger sampling group will be managed.

A ten choice drop-down list has been used in the vocabulary test to measure the performance of the robot. This was done to provide ease and avoid fatigue of the respondents, while simultaneously avoid obtaining a high proportion of outlying responses that could negatively affect the confusion matrix. Possible redesign alternatives in relation to the format of the test within the Participatory Design process essentially fall into one of the following two categories: (1) to have the robot perform more actions, forming complex sentences aiming at completing a full dialogue, or (2) having the respondents provide more custom or personalised answers, moving from a set of closed responses to an open interview format. While these options are not mutually exclusive are definitively appealing, they are prone to lead into the same kind of pitfall, which is: how to quantitatively analyse and evaluate the respondent's answers to obtain statistically relevant results. However, 
there is an incentive for focusing on how to circumvent these challenges, as the long-term goal of this study is to establish a complete and effective human--robot communication. 

Even though there are some potential limitations that need to be handled, such as the need for sign language expertise and the development of a more complex sign language reproduction by TEO, the excellent results obtained with simulation show the importance of focusing on making further advances towards full communication via sign language. One of the considered paths to face these issues is to develop machine learning algorithms to learn from LSE datasets that contain collections of signs performed by professional interpreters. The developed system would additionally enable learning new signs --or in different languages-- from data obtained by low-cost sensors.

\section{Conclusions}

Given the worldwide need for user accessibility and UD in assistive robotics, this work provided a pioneer study of end-user interest, comprehension and satisfaction regarding the reproduction of sign language by a humanoid robot.

The willingness of the end-users of the study towards using sign language with a humanoid robot was almost 94\% positive, which is reaffirmed in the user satisfaction questionnaires after the comprehension tests, where usefulness and future prospects are valued with the highest marks. Both~dactylology and vocabulary tests resulted in 82\% and 83\% correct answer rate respectively, with a relatively pronounced tendency to acceptance in relation to a younger age. Most errors encountered on dactylology and vocabulary should be mendable by modifying finger joint configuration or pronouncing the movement, so further iterations of experiments could be performed to prove if the confusing signs are fixed. Most users distinguished the robot appearance as its most remarkable inconvenience, which is a reasonable outcome since the robot used for testing is an experimental platform and its appearance is constantly changing. 

The most challenging issue regarding this project has been attempting to reproduce sign language with the lack of facial expressions and other non-manual markers. This circumstance may cause understanding problems to some users and would be a potential barrier regarding the development of more complex communication. Concerning basic instructions communication, the tests have shown a proficient human--robot interaction. 

The data collected over these experiments has provided
quantitative measurements on end-user satisfaction, as well as useful insight regarding user needs. The experimental results shed light towards new improvements and developments to make assistive robotics and CPS more usable for deaf and hearing-impaired users. 

%%%%%%%%%%%%%%%%%%%%%%%%%%%%%%%%%%%%%%%%%%
\vspace{6pt} 

%%%%%%%%%%%%%%%%%%%%%%%%%%%%%%%%%%%%%%%%%%

%%%%%%%%%%%%%%%%%%%%%%%%%%%%%%%%%%%%%%%%%%
\authorcontributions{Conceptualization, J.J.G. and J.G.V.; Data curation, J.J.G.; Formal analysis, J.G.V.; Funding acquisition, C.B.; Investigation, J.J.G. and J.G.V.; Methodology, J.J.G.; Project administration, J.G.V. and C.B.; Resources, C.B.; Software, J.J.G. and J.G.V.; Supervision, J.G.V. and C.B.; Validation, J.J.G.; Visualization, J.G.V.; Writing---original draft, J.J.G.; Writing---review and editing, J.G.V. and C.B.}

%%%%%%%%%%%%%%%%%%%%%%%%%%%%%%%%%%%%%%%%%%
\funding{The research leading to these results has received funding from the RoboCity2030-III-CM project (Robótica aplicada a la mejora de la calidad de vida de los ciudadanos. fase III; S2013/MIT-2748), funded by Programas de Actividades I+D en la Comunidad de Madrid and cofunded by Structural Funds of the EU.}

%%%%%%%%%%%%%%%%%%%%%%%%%%%%%%%%%%%%%%%%%%
\acknowledgments{The authors thank CILSEM (Spanish Sign Language Interpreters of Madrid Association) and Signapuntes Lengua de Signos for their kind collaboration with this project.}

%%%%%%%%%%%%%%%%%%%%%%%%%%%%%%%%%%%%%%%%%%
\conflictsofinterest{The authors declare no conflict of interest.} 

%%%%%%%%%%%%%%%%%%%%%%%%%%%%%%%%%%%%%%%%%%
%% optional
\abbreviations{The following abbreviations are used in this manuscript:\\

\noindent 
\begin{tabular}{@{}ll}
BOE & Official Gazette of the Spanish Government\\
CPS & Cyber--Physical Systems \\
DOF & Degrees of Freedom\\
LSE & Spanish Sign Language\\
PD & Participatory Design\\
TEO & Task Environment Operator\\
UC3M & University Carlos III de Madrid\\
UD & Universal Design 

\end{tabular}}

%%%%%%%%%%%%%%%%%%%%%%%%%%%%%%%%%%%%%%%%%%
%% optional
%\appendixtitles{no} %Leave argument "no" if all appendix headings stay EMPTY (then no dot is printed after "Appendix A"). If the appendix sections contain a heading then change the argument to "yes".
%\appendixsections{multiple} %Leave argument "multiple" if there are multiple sections. Then a counter is printed ("Appendix A"). If there is only one appendix section then change the argument to "one" and no counter is printed ("Appendix").
%\appendix
%\section{}
%\unskip
%\subsection{}
%The appendix is an optional section that can contain details and data supplemental to the main text. For example, explanations of experimental details that would disrupt the flow of the main text, but nonetheless remain crucial to understanding and reproducing the research shown; figures of replicates for experiments of which representative data is shown in the main text can be added here if brief, or as Supplementary data. Mathematical proofs of results not central to the paper can be added as an appendix.

%\section{}
%All appendix sections must be cited in the main text. In the appendixes, Figures, Tables, etc. should be labeled starting with `A', e.g., Figure A1, Figure A2, etc. 


%=====================================
% References, variant A: internal bibliography
%=====================================
\reftitle{References}

% The following MDPI journals use author-date citation: Arts, Econometrics, Economies, Genealogy, Humanities, IJFS, JRFM, Laws, Religions, Risks, Social Sciences. For those journals, please follow the formatting guidelines on http://www.mdpi.com/authors/references
% To cite two works by the same author: \citeauthor{ref-journal-1a} (\citeyear{ref-journal-1a}, \citeyear{ref-journal-1b}). This produces: Whittaker (1967, 1975)
% To cite two works by the same author with specific pages: \citeauthor{ref-journal-3a} (\citeyear{ref-journal-3a}, p. 328; \citeyear{ref-journal-3b}, p.475). This produces: Wong (1999, p. 328; 2000, p. 475)

%=====================================
% References, variant B: external bibliography
%=====================================
%\externalbibliography{yes}
%\bibliography{your_external_BibTeX_file}

%%%%%%%%%%%%%%%%%%%%%%%%%%%%%%%%%%%%%%%%%%
%% optional
%\sampleavailability{\hl{Samples of the tests} are available from the authors.}
%%Is this part (sample availability) necessary, can we delete it?

%% for journal Sci
%\reviewreports{\\
%Reviewer 1 comments and authors’ response\\
%Reviewer 2 comments and authors’ response\\
%Reviewer 3 comments and authors’ response
%}

%%%%%%%%%%%%%%%%%%%%%%%%%%%%%%%%%%%%%%%%%%
\end{document}